\documentclass{article}

\usepackage{docmute}

\usepackage{arxiv}

\usepackage[utf8]{inputenc}
\usepackage[T1]{fontenc}

\usepackage{amsmath}
\usepackage{amssymb}
\usepackage{amsfonts}
\usepackage{amsthm}
\usepackage{bm}
\usepackage{comment}
\usepackage{fancybox}
\usepackage{framed}
\usepackage{color}
\usepackage{graphicx}
\usepackage{multicol}
\usepackage{multirow}
\usepackage{cite}
\usepackage{hyperref}

\hypersetup{
    colorlinks = true,
    urlcolor = blue,
    linkcolor = red,
    citecolor = green
}

\usepackage{algorithm}
\usepackage[noend]{algpseudocode}
\usepackage{algorithmicx}

\expandafter\def\expandafter\UrlBreaks\expandafter{\UrlBreaks
  \do\a\do\b\do\c\do\d\do\e\do\f\do\g\do\h\do\i\do\j%
  \do\k\do\l\do\m\do\n\do\o\do\p\do\q\do\r\do\s\do\t%
  \do\u\do\v\do\w\do\x\do\y\do\z\do\A\do\B\do\C\do\D%
  \do\E\do\F\do\G\do\H\do\I\do\J\do\K\do\L\do\M\do\N%
  \do\O\do\P\do\Q\do\R\do\S\do\T\do\U\do\V\do\W\do\X%
  \do\Y\do\Z}

\DeclareMathOperator*{\argmax}{arg\,max}

\algnewcommand\algorithmicforeach{\textbf{for each}}
\algdef{S}[FOR]{ForEach}[1]{\algorithmicforeach\ #1\ \algorithmicdo}

\algnewcommand\AlgAnd{\textbf{and} }
\algnewcommand\AlgOr{\textbf{or} }
\algnewcommand\AlgContinue{\textbf{Continue}}

\algrenewcommand\textproc{}

\algnewcommand{\Initialize}[1]{
	\State \textbf{Initialize:}
 	\State \hspace*{\algorithmicindent}\parbox[t]{0.8\linewidth}{\raggedright #1}}

\algnewcommand{\LeftComment}[1]{
    \Statex $\triangleright$ #1 \hfill}

\def\BibTeX{{\rm B\kern-.05em{\sc i\kern-.025em b}\kern-.08em
    T\kern-.1667em\lower.7ex\hbox{E}\kern-.125emX}}

\usepackage{CJKutf8}

\newcommand{\engtitle}{A Universal LiDAR SLAM Accelerator System on Low-cost FPGA}
\title{\engtitle}

\author{
    Keisuke Sugiura \\
    Keio University \\
    3-14-1 Hiyoshi, Kohoku-ku, Yokohama, Japan \\
    \texttt{sugiura@arc.ics.keio.ac.jp} \\
    \And
    Hiroki Matsutani \\
    Keio University \\
    3-14-1 Hiyoshi, Kohoku-ku, Yokohama, Japan \\
    \texttt{matutani@arc.ics.keio.ac.jp}
}

\hypersetup{
    pdftitle={\engtitle},
    pdfsubject={cs.AR},
    pdfauthor={Keisuke Sugiura, Hiroki Matsutani},
    pdfkeywords={SLAM, Scan Matching, FPGA}
}

\begin{document}

\maketitle



\begin{abstract}
LiDAR (Light Detection and Ranging) SLAM (Simultaneous Localization and Mapping) serves as a basis for indoor cleaning, navigation, and many other useful applications in both industry and household.
From a series of LiDAR scans, it constructs an accurate, globally consistent model of the environment and estimates a robot position inside it.
SLAM is inherently computationally intensive; it is a challenging problem to realize a fast and reliable SLAM system on mobile robots with a limited processing capability.
To overcome such hurdles, in this paper, we propose a universal, low-power, and resource-efficient accelerator design for 2D LiDAR SLAM targeting resource-limited FPGAs.
As scan matching is at the heart of SLAM, the proposed accelerator consists of dedicated scan matching cores on the programmable logic part, and provides software interfaces to facilitate the use.
Our accelerator can be integrated to various SLAM methods including the ROS (Robot Operating System)-based ones, and users can switch to a different method without modifying and re-synthesizing the logic part.
We integrate the accelerator into three widely-used methods, i.e., scan matching, particle filter, and graph-based SLAM.
We evaluate the design in terms of resource utilization, speed, and quality of output results using real-world datasets.
Experiment results on a Pynq-Z2 board demonstrate that our design accelerates scan matching and loop-closure detection tasks by up to 14.84x and 18.92x, yielding 4.67x, 4.00x, and 4.06x overall performance improvement in the above methods, respectively.
Our design enables the real-time performance while consuming only 2.4W and maintaining accuracy, which is comparable to the software counterparts and even the state-of-the-art methods.



\end{abstract}





\section{Introduction} \label{sec:intro}
\begin{figure}
    \centering
    \includegraphics[keepaspectratio,width=0.55\linewidth]{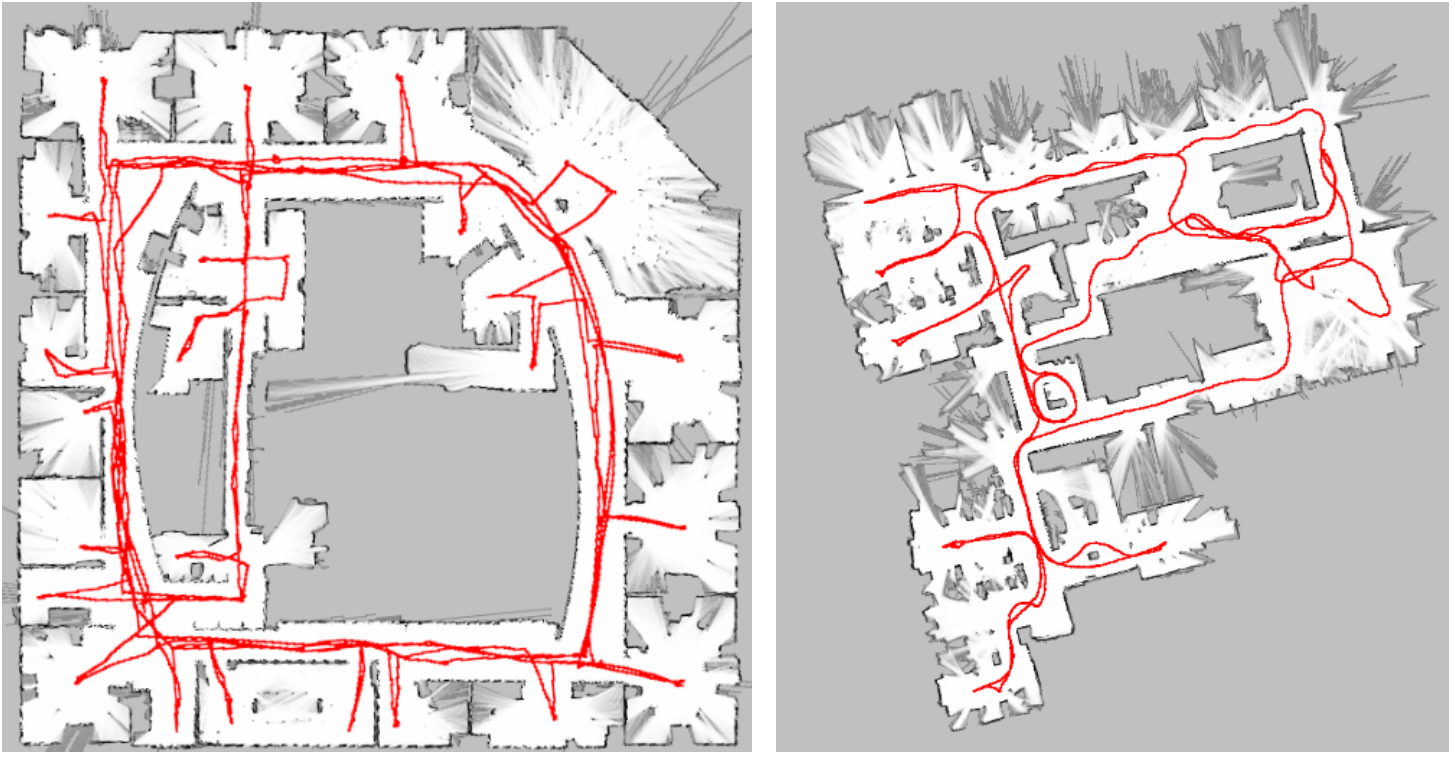}
    \caption{Grid map and robot trajectory (red line) obtained using our proposed FPGA-based SLAM accelerator (left: particle filter-based SLAM and \textbf{Intel} Research Lab dataset, right: graph-based SLAM and \textbf{Revo LDS} dataset)}
    \label{fig:grid-map-intel-revo-lds}
\end{figure}

SLAM (Simultaneous Localization and Mapping) is the task for creating a precise, globally consistent map of a surrounding environment and estimating a robot pose inside the map (Figure \ref{fig:grid-map-intel-revo-lds}).
It is a chicken-and-egg problem, since a correct map is required to accurately estimate a robot pose, and at the same time a correct pose is required to build a map from a sensor input.
It plays a central role for positioning systems on autonomous robots especially in GPS-denied scenarios.
There is a growing demand for autonomous indoor mobile robots with reliable SLAM systems, which have many useful applications such as house cleaning~\cite{Lee13}, indoor navigation~\cite{Atia15,QinZou21}, and transportation in warehouses~\cite{HanWang21,Moura21}.
SLAM systems for such robots should be high-performance and also power-efficient, since onboard computational resources and power supply are limited because of cost and payload constraints.

A number of SLAM methods have been developed for more than two decades, which can be categorized according to the dimensionality of space (2D, 3D), sensor configuration (LiDAR, camera), and underlying mathematical formulation (filtering-based, optimization-based).
LiDAR (Light Detection and Ranging) and camera are the two commonly-used sensors in SLAM.
LiDAR has certain advantages over cameras: it provides highly-accurate spatial information about the environment in the form of point clouds, which can be immediately used in a SLAM process, whereas geometrical information should be first extracted from a series of images, incurring additional computational complexity.
Unlike cameras, LiDAR is not affected by illumination changes and has a wider field-of-view.
LiDAR-based SLAM ~\cite{JiZhang14,JiZhang15,Hess16,TixiaoShan18,TixiaoShan20,CorneliaSchulz20,HanWang21} achieves high accuracy owing to the above characteristics, and outperforms vision-based SLAM in famous datasets (e.g., KITTI odometry~\cite{Geiger12}).
However, LiDAR SLAM methods usually require desktop or laptop computers equipped with high-end CPUs to deal with high computational loads and achieve satisfactory performance~\cite{Hess16,TixiaoShan20,CorneliaSchulz20,HanWang21}.
Additionally, their application to the low-power computing platform has yet to be explored despite of its importance.
An FPGA-based LiDAR SLAM system that achieves energy efficiency and real-time performance is the promising solution to tackle this problem, and should bring significant benefits to both industry and household.

Scan matching lies at the core in LiDAR SLAM algorithms: it is the process of estimating a relative robot motion in the form of 2D or 3D rigid transformation by registering a new LiDAR scan to a reference map, or aligning consecutive LiDAR scans, based on which a current robot pose is incrementally updated.
The method should be carefully chosen, since scan matching is a performance bottleneck and has a great influence on the quality of outputs in many cases.
Scan matching is a mature and well-established technique; ICP (Iterative Closest Point) with its variants~\cite{Besl92,Censi08,Segal09,JiaolongYang16} and NDT (Normal Distributions Transform)~\cite{Biber03} are the famous approaches, while other methods leverage histogram-based feature descriptors~\cite{Rusu08,Rusu09}, deep learning techniques~\cite{Elbaz17,JiaxinLi17,Aoki19,Yuewang19}, correlation between scan and map~\cite{Konolige99,Olson09,Chong13}, and branch-and-bound~\cite{Olson15,Hess16,Daun19}.
FPGA-based accelerator for robust scan matching is essential for realizing a reliable and power-efficient LiDAR SLAM on low-cost indoor mobile robots.
It reduces power consumption and also CPU time spent for SLAM process: robots can use that time for other purposes such as path planning and navigation.

2D LiDAR SLAM takes a sequence of scans from a 2D LiDAR as input, and produces a sparse map consisting of salient feature points (i.e., landmarks) or a dense occupancy grid map.
The suitable method depends on the environmental conditions, performance requirements, and LiDAR sensor specifications.
For instance, particle filter-based~\cite{Grisetti05,Grisetti07A,Grisetti07B} and graph-based~\cite{Grisetti10,Konolige10,Kuemmerle11} methods perform loop-closure detection, i.e., checks whether a robot is revisiting already explored areas to eliminate the accumulated error, and thereby can be applied for large indoor environments.
On the other hand, for a small structured environments with distinct features, such loop detection is not always required; a simpler approach based on sequential scan matching~\cite{Kohlbrecher11} is adequate for building a correct map.
Considering the above, an FPGA-based accelerator would be more beneficial if it is compatible with a variety of LiDAR SLAM methods, rather than supporting only one specific method.

In this paper, we propose a unified accelerator design for 2D LiDAR SLAM methods targeting low-cost FPGA SoC (Figure \ref{fig:design-block-diagram}).
Our accelerator implements a scan-to-map matching, i.e., it aligns an input LiDAR scan with a dense occupancy grid map and computes an optimal sensor pose with respect to the map.
As scan-to-map matching is generally more robust than scan-to-scan matching, our accelerator enables an FPGA-based SLAM system with a sufficient accuracy.
To our knowledge, this is an initial work that consider an FPGA-based hardware accelerator for LiDAR SLAM.
The novelty and main contributions in this paper are summarized as follows:
\begin{enumerate}
  \item We propose a design which is modular in a way that it can be integrated to various LiDAR SLAM methods.
This allows users to choose the most suitable one according to performance requirements and environmental characteristics.
The proposed resource-efficient design is implemented on Pynq-Z2 board, and applied to three representative LiDAR SLAM approaches: scan matching-based, particle filter-based, and graph-based SLAM.
  \item To maximize the overall performance, we implement the scan matching part on the FPGA fabric.
Specifically, we develop an accelerator for Correlative Scan Matching (CSM) proposed by Olson \textit{et al.}~\cite{Olson09}, termed CSM core, since it strikes a balance between computational efficiency and robustness.
We also modify the CSM algorithm to derive a (partially) parallelized and hardware-friendly version.
We then optimize the design by simple data reordering and loop interchanging.
Our design consists of two CSM cores, which can be used for different purposes (e.g., scan matching and loop detection) at the same time.
  \item The proposed FPGA accelerator is ROS (Robot Operating System)-compliant, and supports the ROS-based SLAM systems.
More specifically, we provide an interface to handle communications between FPGA fabric and ROS framework.
  \item As a scan matching-based SLAM, we consider Hector SLAM proposed by Kohlbrecher \textit{et al.}~\cite{Kohlbrecher11}.
  We propose a robustified version of it by exploiting the advantages of CSM.
  We then integrate the accelerator into the ROS-based implementation for better performance using the above ROS-FPGA interface.
  \item Experimental results using real-world datasets confirm that the performance is effectively boosted while preserving accuracy, and highlight the trade-offs between computational cost and accuracy in these methods, which would support our unified design concept.
\end{enumerate}

This paper is outlined as follows: the next section presents related works for FPGA-based SLAM accelerators.
Section \ref{sec:prelim} is devoted to a brief explanation of the three SLAM approaches mentioned above along with scan matching.
Various design optimizations are described in Section \ref{sec:design}.
Section \ref{sec:eval} illustrates the implementation details and shows the evaluation results in terms of resource utilization, power consumption, throughput, and quality of outputs.
Section \ref{sec:conc} concludes this paper.


\section{Related Works} \label{sec:related}

A number of visual SLAM systems for an FPGA SoC platform have been proposed in the literature~\cite{JanoschNikolic14,QuentinGautier14,KonstantinosBoikos16,KonstantinosBoikos17,WeikangFang17,QuentinGautier19,RunzeLiu19,ZhilinXu20,ChengWang21,Gkeka21}.
The time-consuming and computationally intensive parts such as corner detection~\cite{JanoschNikolic14}, feature extraction~\cite{WeikangFang17,RunzeLiu19,ZhilinXu20,ChengWang21}, and tracking~\cite{QuentinGautier14,KonstantinosBoikos16,KonstantinosBoikos17} are implemented on the programmable logic part, showing that FPGA enables low-cost yet high-performance SLAM systems.

Boikos \textit{et al.}~\cite{KonstantinosBoikos17} proposed an accelerator for direct photometric tracking in LSD-SLAM, which is the process of estimating a camera pose by minimizing per-pixel intensity differences between two images.
Their implementation using Zedboard (Xilinx Zynq-7020 SoC) achieved 22.7fps at an input resolution of 320x240, which is 10x faster than a dual-core ARM Cortex-A9 CPU.
Gautier \textit{et al.}~\cite{QuentinGautier19} proposed an algorithm that combines depth fusion and ray casting steps in InfiniTAM, and its implementation on a Terasic DE1 SoC (Altera Cyclone V) improved the overall performance by 3.16x (0.49 to 1.55fps, at a 320x180 resolution) compared to ARM Cortex-A9.
Liu \textit{et al.}~\cite{RunzeLiu19} devised a rotationally-invariant feature descriptor for a hardware-friendly implementation of feature detection and matching in ORB-SLAM.
Their proposed system using Xilinx Zynq XCZ7045 SoC (Kintex 7) gave 31x better throughput than ARM Cortex-A9 at a 640x480 resolution.

Xu \textit{et al.}~\cite{ZhilinXu20} implemented SuperPoint, which is a CNN-based feature extraction method, on the Xilinx ZCU102 evaluation board.
More specifically, they use a custom accelerator for post-processing operations (e.g., normalization and softmax) in conjunction with Xilinx DPU (DNN Processing Unit) for offloading the CNN backbone of SuperPoint, both of which exploit fixed-point quantizations.
Wang \textit{et al.}~\cite{ChengWang21} proposed ac$^2$SLAM, a hardware optimized version of ORB-SLAM2.
The design consists of a ping-pong heapsort module and a scalable ORB extractor for keypoint selection and detection.
Their implementation on Xilinx ZCU104 evaluation kit yielded a 40x speedup for feature extraction and a 2.7x framerate improvement compared to a quad-core ARM Cortex-A53.

Compared to visual SLAM, only a few studies have considered an FPGA acceleration of LiDAR SLAM despite its importance and widespread use.
In \cite{HaoSun20}, a novel voxel-based data structure for hierarchical point cloud partitioning is proposed, and an FPGA-based KNN (k-nearest neighbor) accelerator for 3D point cloud matching is also presented.
Though the accelerator provides orders of magnitude faster search speed than CPU counterpart while consuming less energy, its performance effects on 3D LiDAR SLAM are not evaluated.
Additionally, our design adopts a scan-to-map matching method, which does not require the expensive correspondence search between point clouds by making use of an occupancy grid map representation.

In \cite{JoshuaFrank21}, the authors proposed an approach similar to this paper.
Since CSM becomes a bottleneck part in a state-of-the-art 2D LiDAR SLAM, Google Cartographer~\cite{Hess16}, they devised an FPGA-based accelerator for CSM using Xilinx ZCU104, that evaluates correspondence scores (i.e., degree of overlap between LiDAR scan and occupancy grid map) for given candidate solutions.
Rather than focusing on a score evaluation part, we develop a custom IP core (CSM core) that performs the entire CSM algorithm, and our board-level design fits within a low-cost FPGA (Pynq-Z2).
In addition, the CSM core parallelizes the score evaluation for multiple candidate solutions.

The authors of \cite{QiDeng21} proposed an implementation for semantic-assisted NDT-based scan registration on Xilinx ZCU102, and achieved a 35.85x speedup and a 14.3x better energy efficiency compared to ARM Cortex-A53.
NDT optimizes a rigid transformation between LiDAR scan and map on a continuous search space in an iterative fashion, and requires a computation of probability density for all points in a scan.
Compared to that, the scan matching method we used finds an optimal transformation on a discrete search space (i.e., on a finite set of candidate solutions), reducing the fixed-point operations to some extent and avoids the suboptimal solutions.

In \cite{Eisoldt21} and \cite{Flottmann21}, the authors presented a LiDAR SLAM system mounted with a LiDAR that performs iterative scan-to-map registration and updates a 3D TSDF (Truncated Signed Distance Functions) map on the FPGA fabric.
They implemented the design using a Trenz UltraSom+ module (Xilinx UltraScale+ MPSoC), resulting in a 7x throughput improvement and 18x energy consumption reduction compared to Intel NUC (Intel Core i7-6770HQ).
Their SLAM systems also provide a ROS interface for transferring sensor data and inspecting the mapping process.
While these works have successfully demonstrated a 3D LiDAR SLAM system on a mid-range FPGA, we focus on a more compact, resource-efficient design for 2D SLAM, and perform several optimizations in both algorithmic and architectural aspects.

\section{Preliminaries} \label{sec:prelim}
We firstly overview the 2D scan matching based on LiDAR sensors and occupancy grid maps.
Then, we describe the three representative categories of 2D LiDAR SLAM, namely scan matching-based, particle filter-based, and graph-based SLAM in the following sections.

\subsection{Scan Matching-based SLAM} \label{sec:prelim-scan-matching-slam}
\begin{figure}
    \centering
    \includegraphics[keepaspectratio,width=0.55\linewidth]{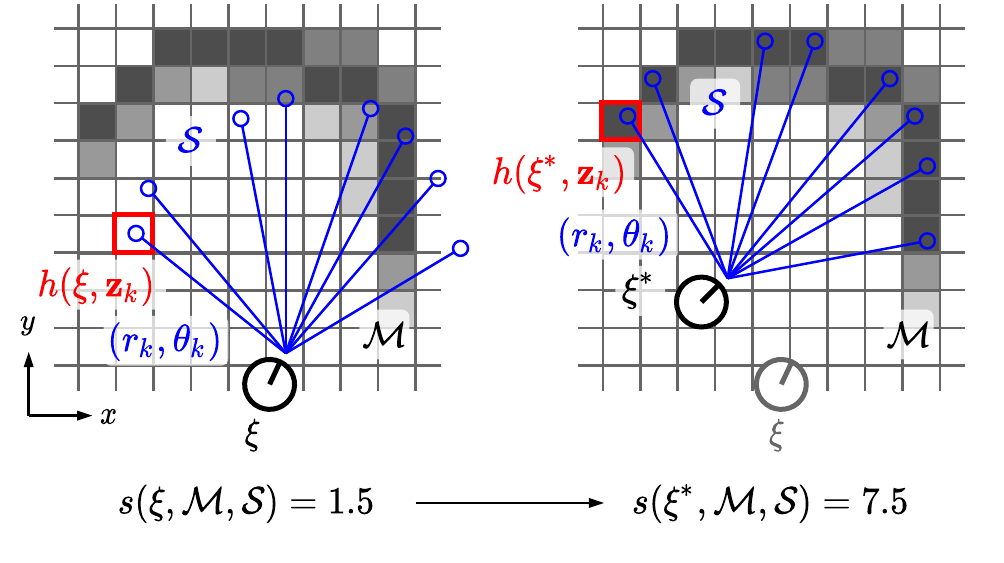}
    \caption{Updating a robot pose $\bm{\xi}$ by scan matching.
    Blue circle dots represent scan points projected onto a grid map $\mathcal{M}$, and squares filled with darker color indicate grid cells with higher occupancy probabilities.
    Scan matching aligns a scan $\mathcal{S}$ with a map $\mathcal{M}$ and seeks a robot pose $\bm{\xi}^*$ (right), by maximizing a score $s(\bm{\xi}; \mathcal{M}, \mathcal{S})$.
    The score is obtained by summing up occupancy probabilities of grid cells which correspond to scan points.}
    \label{fig:map-and-scan}
\end{figure}

As shown in Figure \ref{fig:map-and-scan}, scan matching is the process of aligning a LiDAR scan $\mathcal{S}$ with an occupancy grid map $\mathcal{M}$ and computing a pose of the scan $\bm{\xi}^*$, from which a relative robot motion is obtained.
It is the fundamental and most important part in SLAM, and directly affects the quality of outputs.

The scan $\mathcal{S} = \left\{ \mathbf{z}_1, \ldots, \mathbf{z}_N \right\}$ forms a point cloud representing obstacles around a LiDAR sensor ($N$ is the number of points).
Each scan point $\mathbf{z}_i = (r_i, \theta_i)$ is defined by a range $r_i$ and an angle $\theta_i$ in the polar coordinate frame centered at the LiDAR sensor.
The grid map $\mathcal{M}$ partitions the surrounding environment into equally-sized grid cells with a resolution of $r$, each of which stores an occupancy probability that the cell is occupied by an obstacle.
We denote the occupancy probability at a grid cell $(i, j) \in \mathbb{N}^2$ as $\mathcal{M}(i, j) \in \left[ 0, 1 \right]$.
Using these notations, the scan matching problem is formalized as a maximization of a score $s(\bm{\xi}; \mathcal{M}, \mathcal{S})$ with respect to a scan pose $\bm{\xi}$ as follows:
\begin{equation}
    \bm{\xi}^* = \argmax_{\bm{\xi}} s(\bm{\xi}; \mathcal{M}, \mathcal{S}).
    \label{eq:scan-matching-problem}
\end{equation}
The score correlates to the quality of an alignment between map and scan under the current pose estimate $\bm{\xi}$.
We compute the score by projecting scan points onto grid cells and summing up their associated occupancy probabilities:
\begin{equation}
    s(\bm{\xi}, \mathcal{M}, \mathcal{S})
    = \sum_k \mathcal{M} \left( h(\bm{\xi}, \mathbf{z}_k) \right),
    \label{eq:matching-score}
\end{equation}
where $h(\bm{\xi}, \mathbf{z}_k)$ converts the $k$-th scan point $\mathbf{z}_k = (r_k, \theta_k)$ to its corresponding grid cell indices given a pose $\bm{\xi} = \left[ \xi_x, \xi_y, \xi_\theta \right]^\top$:
\begin{equation}
    h(\bm{\xi}, \mathbf{z}_k) = \left(
    \left\lfloor \frac{r_k \cos(\theta_k + \xi_\theta) + \xi_x}{r} \right\rfloor, \
    \left\lfloor \frac{r_k \sin(\theta_k + \xi_\theta) + \xi_y}{r} \right\rfloor \right).
    \label{eq:scan-point-to-map}
\end{equation}
If a scan is sufficiently aligned with a map, an occupancy probability $\mathcal{M}(h(\bm{\xi}, \mathbf{z}_i))$ for each scan endpoint $\mathbf{z}_i$ is closer to one since it represents the presence of an obstacle, thus resulting in a higher score.
Scan matching-based SLAM~\cite{Biber03,Kohlbrecher11} is a simple form of SLAM and as its name suggests, it involves a sequential scan matching, i.e., it updates a current robot pose $\bm{\xi}_t$ by matching the latest scan $\mathcal{S}_t$ with a grid map consisting of previously acquired scans.

\subsection{Particle Filter-based SLAM (PF-SLAM)} \label{sec:prelim-pf-slam}
PF-SLAM~\cite{Grisetti05,Grisetti07A,Grisetti07B} uses a set of particles, each of which carries a hypothesis about a robot trajectory $\left\{ \bm{\xi}_0, \ldots, \bm{\xi}_t \right\}$ and a map $\mathcal{M}$.
To maintain plausible hypotheses, particles with larger importance weights are replicated and propagated to the next time step, while ones with smaller weights are removed.
Importance weight indicates the consistency of particle's map and current pose $(\mathcal{M}_i, \bm{\xi}_i)$ with the latest scan input $\mathcal{S}_t$, which is computed from the score $s(\bm{\xi}_i; \mathcal{M}_i, \mathcal{S})$ in the scan matching process.
PF-SLAM improves reliability by employing multiple particles at the cost of increased computational complexity, as scan matching is performed for each particle.
Scan matching is the most suitable candidate for hardware acceleration, as the operation for each particle is independent and completely parallelizable~\cite{Gouveia14,Sileshi16A,Sileshi16B,Krishna21}.

\subsection{Graph-based SLAM} \label{sec:graph-slam}
Graph-based SLAM~\cite{Grisetti10,Konolige10,Kuemmerle11} takes a different approach for estimating a robot trajectory: it creates a pose graph, whose node represents a pose $\bm{\xi}_t$ at time $t$ and edge $(\bm{\Delta \xi}_{i, j}, \bm{\Sigma}_{i, j}^{-1})$ defines a relative pose $\bm{\Delta \xi}_{i, j}$ between two nodes $i, j$ along with its uncertainty $\bm{\Sigma}_{i, j}^{-1}$.
Edges represent either odometry or loop: the former is incrementally created in the frontend scan matching, where a new scan is aligned with the latest submap to estimate the current robot pose (submap is created from a series of scans received within a certain time window).
The latter is inserted after a successful loop detection in the backend, which involves scan matching between recent scans and old submaps to detect whether a robot is revisiting previously explored areas.
Backend also optimizes the pose graph, i.e., reduces inconsistencies between nodes and edges, to refine the entire trajectory estimate.

Instead of tracking multiple hypotheses like PF-SLAM, graph-based SLAM only maintains the most plausible estimate of trajectory and map, and improves accuracy by explicitly performing loop detections.
This contributes to the reduced computational cost compared to PF-SLAM.
As explained above, both frontend and backend rely heavily on scan matching, which motivates the hardware acceleration of scan matching in graph-based SLAM~\cite{DonghwaLee12,Ratter13,Ratter15} as well as PF-SLAM and scan matching-based SLAM.

\subsection{Correlative Scan Matching (CSM)} \label{sec:slam-csm}
Using Equation \ref{eq:scan-matching-problem}, scan matching repetitively evaluates scores for a large number of candidate solutions.
This involves memory accesses to grid cells and coordinate transformations (as seen in Equation \ref{eq:scan-point-to-map}) for scan points.
Scan matching algorithm makes no assumption for the shapes of input scans, meaning that memory access patterns for grid maps are irregular and unpredictable as shown in Figure \ref{fig:map-and-scan}, reducing the cache performance.
It is therefore both compute- and data-intensive process and inevitably becomes a major bottleneck in SLAM applications.
We aim to address this by FPGA-based scan matching acceleration and enable SLAM on low-power edge devices.
In the following, we describe the Correlative Scan Matching (CSM) algorithm~\cite{Olson09} and highlight its advantages in terms of hardware efficiency by comparing it to other existing methods.

Algorithm \ref{alg:csm} summarizes the CSM algorithm~\cite{Olson09}.
Let $\mathcal{W} = \left[ -w_x, w_x \right) \times \left[ -w_y, w_y \right) \times \left[ -w_\theta, w_\theta \right)$ be a discrete search window of size $(2w_x, 2w_y, 2w_\theta)$, and $(n_x, n_y, n_\theta) \in \mathcal{W}$ be a candidate solution inside the window.
CSM finds the optimal solution $(n_x^*, n_y^*, n_\theta^*) \in \mathcal{W}$ with the maximum score $s(n_x^*, n_y^*, n_\theta^*)$, from which the pose estimate $\bm{\xi}^* = \left[ \xi_x^0 + r n_x^*, \xi_y^0 + r n_y^*, \xi_\theta^0 + \delta_\theta n_\theta^* \right]$ is obtained (line \ref{alg:csm-final-pose}).
Note that $r, \delta_\theta$ denote step sizes along $x, y$ and $\theta$ directions, respectively.
$\bm{\xi}^0 = \left[ \xi_x^0, \xi_y^0, \xi_\theta^0 \right]$ is the pose at the search window center (i.e., $n_x, n_y, n_\theta = 0$).

\begin{figure}
    \centering
    \includegraphics[keepaspectratio,width=0.55\linewidth]{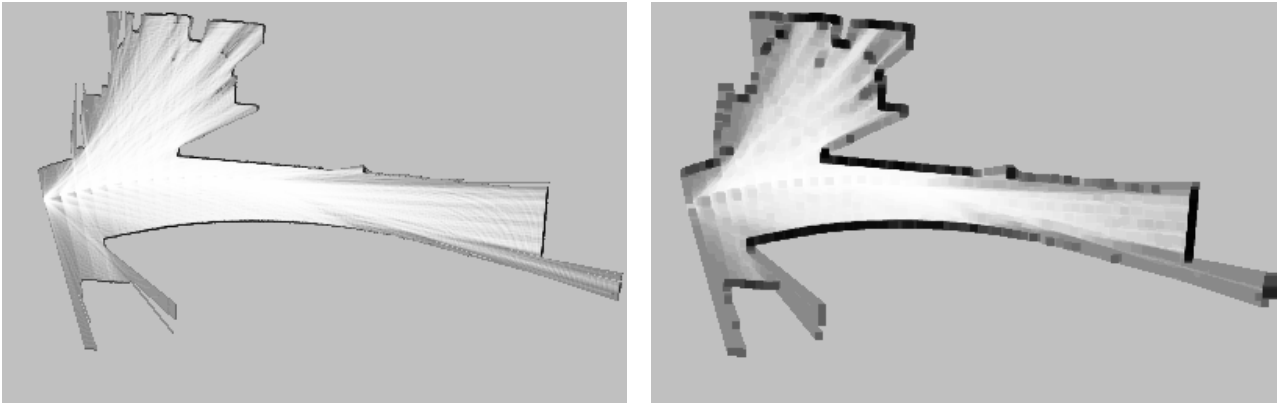}
    \caption{Example of \textbf{fine} $\mathcal{M}$ (left) and \textbf{coarse} $\mathcal{M}'$ (right) grid maps}
    \label{fig:fine-and-coarse-map}
\end{figure}

The algorithm is basically a special case of Branch-and-Bound (BB)~\cite{Olson15,Hess16} and utilizes two-level resolution grid maps (i.e., coarse map $\mathcal{M}'$ and fine map $\mathcal{M}$, Figure \ref{fig:fine-and-coarse-map}).
The coarse map $\mathcal{M}'$ is computed from the input map $\mathcal{M}$ using the \textbf{sliding window maximum} operation (line \ref{alg:csm-coarse-map}) as follows:
\begin{equation}
    \mathcal{M}'(i, j) = \max_{i', j' \in \left[ 0, 1, \ldots, w - 1 \right]}
    \mathcal{M}(i + i', j + j').
    \label{eq:csm-sliding-window-maximum}
\end{equation}
Each cell in $\mathcal{M}'$ contains a local maximum within a corresponding small region (of $w \times w$ cells) in $\mathcal{M}$.

The outermost loop (line \ref{alg:csm-loop-theta}) iterates over $\theta$.
For each angle $n_\theta \in \left[ -w_\theta, w_\theta \right)$, CSM projects scan points onto grid cells and computes a set of indices $\mathcal{I} = \left\{ (i_1, j_1), \ldots, (i_N, j_N) \right\}$, where $(i_k, j_k) = h(\bm{\xi}, \bm{z}_k)$ is the index corresponding to the $k$-th scan point, and $\bm{\xi} = \left[ \xi_x^0, \xi_y^0, \xi_\theta^0 + \delta_\theta n_\theta \right]$ is the pose for a candidate solution $(0, 0, n_\theta) \in \mathcal{W}$.
Using $\mathcal{I}$, we can compute a score for a candidate solution $(n_x, n_y, n_\theta)$ as follows:
\begin{equation}
    s(n_x, n_y, n_\theta) = \sum_k \mathcal{M}(i_k + n_x, j_k + n_y).
    \label{eq:csm-evaluate-fine}
\end{equation}

CSM takes a \textbf{coarse-to-fine} matching strategy.
In \textbf{coarse matching} step (lines \ref{alg:csm-coarse-loop-y}-\ref{alg:csm-coarse-eval-continue}), CSM evaluates scores $s'$ for all points ($n_x', n_y', n_\theta$) in a coarse search grid using a coarse map $\mathcal{M}'$ as follows:
\begin{equation}
    s'(n_x', n_y', n_\theta) = \sum_k \mathcal{M}'(i_k + n_x', j_k + n_y').
    \label{eq:csm-evaluate-coarse}
\end{equation}
By substituting Equations \ref{eq:csm-sliding-window-maximum}-\ref{eq:csm-evaluate-fine} into \ref{eq:csm-evaluate-coarse}, we obtain:
\begin{eqnarray}
    s' &=& \sum_k \max_{i', j' \in \left[ 0, 1, \ldots, w - 1 \right]}
    \mathcal{M}(i_k + n_x' + i', j_k + n_y' + j') \nonumber \\
    &\ge& \max_{i', j' \in \left[ 0, 1, \ldots, w - 1 \right]}
    \sum_k \mathcal{M}(i_k + n_x' + i', j_k + n_y' + j') \nonumber \\
    &=& \max_{i', j' \in \left[ 0, 1, \ldots, w - 1 \right]} s(n_x' + i', n_y' + j', n_\theta).
\end{eqnarray}
This inequality ensures that the score $s'(n_x', n_y', n_\theta)$ evaluated using Equation \ref{eq:csm-evaluate-coarse} is the upper-bound of scores in the $w \times w$ search space starting from $(n_x', n_y')$.
This allows the pruning at line \ref{alg:csm-coarse-eval-continue} and improves the algorithm efficiency.
As a result, CSM is able to identify the region of interest that contains a global optimum $\bm{\xi}^*$, and rule out the large part of the search space.

In \textbf{fine matching} step (lines \ref{alg:csm-fine-loop-y}-\ref{alg:csm-solution-update}), a score $s$ for each point $(n_x, n_y, n_\theta)$ in a $w \times w$ region spanning from $(n_x', n_y', n_\theta)$ to $(n_x' + w - 1, n_y' + w - 1, n_\theta)$ is evaluated using the fine map $\mathcal{M}$ and Equation \ref{eq:csm-evaluate-fine}.
The current solution $(n_x^*, n_y^*, n_\theta^*), s^*$ is updated as necessary (line \ref{alg:csm-solution-update}).

Throughout the paper, we set to $r = 5 \mathrm{cm}$ and $w = 8$.
In this setting, coarse matching sweeps the entire search space at $r \cdot w = 40 \mathrm{cm}$ resolution, and fine matching tries to find the optimal solution from $w^2 = 64$ candidate points equally spaced $5 \mathrm{cm}$ apart inside the area of size $40 \times 40 \mathrm{cm}$.
As apparent in line \ref{alg:csm-final-pose}, CSM estimates translational and rotational components of the pose $\bm{\xi}^*$ at $r = 5 \mathrm{cm}$ and $\delta_\theta \ \mathrm{rad}$ accuracy.


\begin{algorithm}[h]
  \caption{Correlative Scan Matching (CSM)}
  \label{alg:csm}
  \begin{algorithmic}[1]
    \Require $\mathcal{M}$, $\mathcal{S}$, $\bm{\xi}^0$, $(w_x, w_y, w_\theta)$, $r$, $\delta_\theta$, $w$
    \Ensure Optimal pose $\bm{\xi}^*$, score $s^*$
    \State Compute $\widehat{w}_x, \widehat{w}_y$ such that $2 w_x = w \cdot \widehat{w}_x, 2 w_y = w \cdot \widehat{w}_y$
    \State Compute a coarse map $\mathcal{M}'$ using (\ref{eq:csm-sliding-window-maximum})
    \label{alg:csm-coarse-map}
    \State $s^* \gets -\infty, (n_x^*, n_y^*, n_\theta^*) \gets (-w_x, -w_y, -w_\theta)$
    \For{$n_\theta = -w_\theta, \ldots, w_\theta - 1$} \label{alg:csm-loop-theta}
      \LeftComment{\textbf{Scan discretization}}
      \State Compute a set of indices $\mathcal{I} = \left\{ (i_k, j_k) \right\}$ using (\ref{eq:scan-point-to-map}) \label{alg:csm-discretize}
      \LeftComment{\textbf{Coarse matching}}
      \For{$\widehat{n}_y = 0, 1, \ldots, \widehat{w}_y - 1$} \label{alg:csm-coarse-loop-y}
        \For{$\widehat{n}_x = 0, 1, \ldots, \widehat{w}_x - 1$} \label{alg:csm-coarse-loop-x}
          \State $n_x' \gets -w_x + \widehat{n}_x \cdot w, \ n_y' \gets -w_y + \widehat{n}_y \cdot w$
          \State $s' \gets \sum_{k = 1}^N \mathcal{M}'(i_k + n_x', j_k + n_y')$
          (\ref{eq:csm-evaluate-coarse}) \label{alg:csm-coarse-eval}
          \If{$s' \le s^*$}
            \State \AlgContinue \label{alg:csm-coarse-eval-continue}
          \EndIf
          \LeftComment{\textbf{Fine matching}}
          \For{$n_y = n_y', \ldots, n_y' + w - 1$} \label{alg:csm-fine-loop-y}
            \For{$n_x = n_x', \ldots, n_x' + w - 1$} \label{alg:csm-fine-loop-x}
              \State $s \gets \sum_{k = 1}^N \mathcal{M}(i_k + n_x, j_k + n_y)
              (\ref{eq:csm-evaluate-fine}) $\label{alg:csm-fine-eval}
              \If{$s > s^*$}
                \State $s^* \gets s, (n_x^*, n_y^*, n_\theta^*) \gets (n_x, n_y, n_\theta)$
                \label{alg:csm-solution-update}
              \EndIf
            \EndFor
          \EndFor
        \EndFor
      \EndFor
    \EndFor
    \State \Return $\bm{\xi}^* = \left[ \xi_x^0 + r n_x^*, \xi_y^0 + r n_y^*, \
    \xi_\theta^0 + \delta_\theta n_\theta^* \right], s^*$ \label{alg:csm-final-pose}
  \end{algorithmic}
\end{algorithm}


As shown above, CSM is the simplified version of BB and is classified as a non-iterative method.
Unlike BB counterparts~\cite{Olson15,Hess16,Daun19}, which use multi-resolution maps and tree data structures, CSM~\cite{Olson09} only requires a precomputation of a map $\mathcal{M}'$ and works without any complex data structure as evident in Algorithm \ref{alg:csm}.
This reduces the preprocessing overhead and resource consumptions in exchange for a slight efficiency loss.
Compared to iterative methods like hill-climbing~\cite{Montemerlo03B} or Gauss-Newton (Levenberg-Marquardt) method~\cite{Biber03,Kohlbrecher11,Fossel15,Pedrosa16,Pedrosa17}, which tend to be trapped in local optima, CSM does not depend on initial guesses and guarantees the globally-optimal solution in the predefined search window.
From the above considerations, we choose to use CSM in our implementation, since it is robust, hardware-friendly, and strikes a balance between algorithm simplicity and efficiency.
We describe the design and implementation of our FPGA-based CSM core in the next section.

\section{Design Optimization} \label{sec:design}
\subsection{Overview of the CSM Core Design} \label{sec:design-overview}
\begin{figure}[htbp]
    \centering
    \includegraphics[keepaspectratio,width=0.55\linewidth]{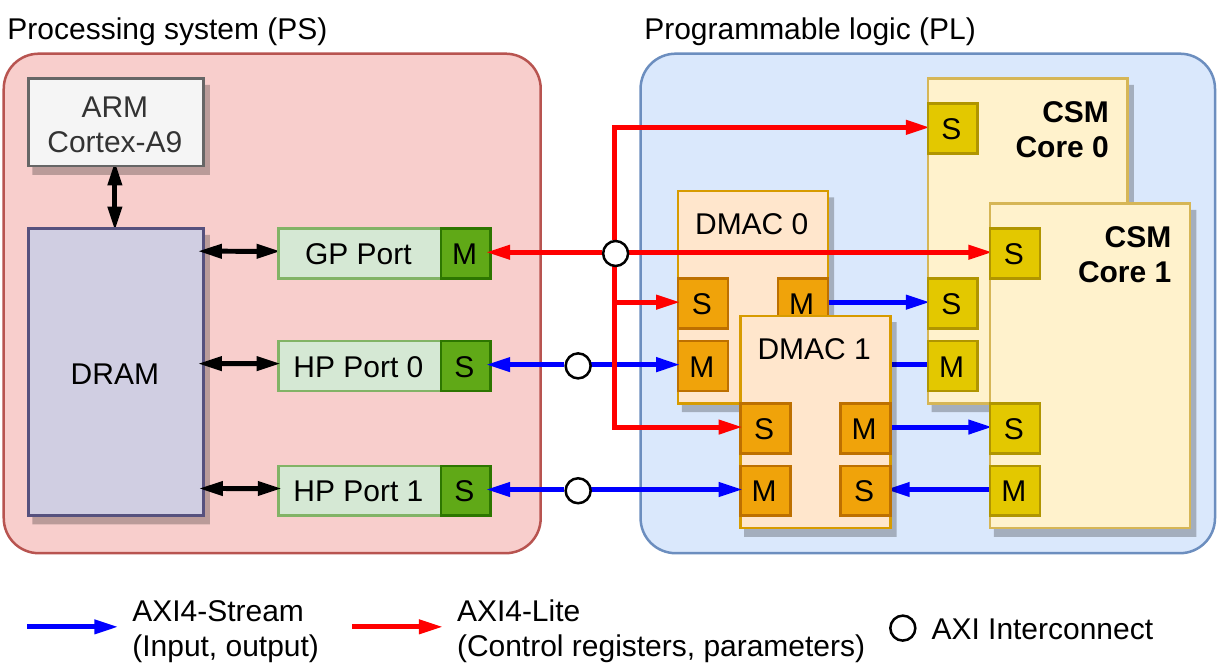}
    \caption{Block diagram of the implementation}
    \label{fig:design-block-diagram}
\end{figure}

Figure \ref{fig:design-block-diagram} illustrates the block diagram of the board-level implementation.
The description below assumes Xilinx Zynq SoC devices, which is composed of a programmable logic (PL) part and a processing system (PS) part.
The PL part contains two CSM cores, each of which independently performs CSM algorithm in response to the request from the PS part.
CSM is performed by the following steps with a help of the CSM IP core interface:
\begin{enumerate}
    \item The IP core interface retrieves a scan matching query containing an input data, i.e., a grid map $\mathcal{M}$, a LiDAR scan $\mathcal{S}$, and an initial pose $\bm{\xi}^0$ from an upstream SLAM module.
    It places the scan $\mathcal{S}$ and the map $\mathcal{M}$ into the contiguous DRAM buffer.
    As discussed in Section \ref{sec:design-opt-buffer-size}, only a portion of the map visible from the current LiDAR position is written to the buffer.
    \item The interface then transfers the input data to the CSM core via a high-speed AXI4-Stream interface by triggering a DMA (Direct Memory Access) controller.
    It also sets the algorithmic parameters of CSM through memory-mapped I/O and starts the CSM core.
    \item It waits until the results are ready by polling a status register of the DMA controller.
    After the CSM core completes its job, the interface receives a resulting pose $\bm{\xi}^*$ and a matching score $s^*$ via DMA and passes them to the upstream SLAM module.
\end{enumerate}
Importantly, the design is independent from the choice of 2D LiDAR SLAM methods, and users can switch to different suitable methods (e.g., PF-SLAM, graph-based SLAM, and Hector SLAM) to adapt to the environments and performance requirements without modifying and re-synthesizing the design.
Our design also incorporates the CSM core interfaces as described in Sections \ref{sec:design-application-to-slam} and \ref{sec:design-application-to-hector-slam}, which hide unnecessary details regarding to memory-mapped I/O and DMA transfers as above and facilitates the use of CSM cores.

Our design conforms to the AXI4-Stream protocol and uses one 64-bit high-performance (HP) port for each DMA controller for the high-speed burst transfer between DRAM and on-chip block RAM (BRAM) (blue arrows in Figure \ref{fig:design-block-diagram}).
The control registers are accessible via an AXI4-Lite interface connected to a GP port (red arrows in Figure \ref{fig:design-block-diagram}).
PS writes to the registers in DMA controllers to set the length and physical address of the data to be transferred.
CSM core also provides a set of registers to specify the algorithmic parameters such as grid map size, search window size $(w_x, w_y, w_\theta)$, search step $(r, \delta_\theta)$, and number of points in a scan $\mathcal{S}$ (see Algorithm \ref{alg:csm}).

\subsection{Details of the CSM Core Design} \label{sec:design-details}
\begin{figure}[htbp]
    \centering
    \includegraphics[keepaspectratio,width=0.55\linewidth]{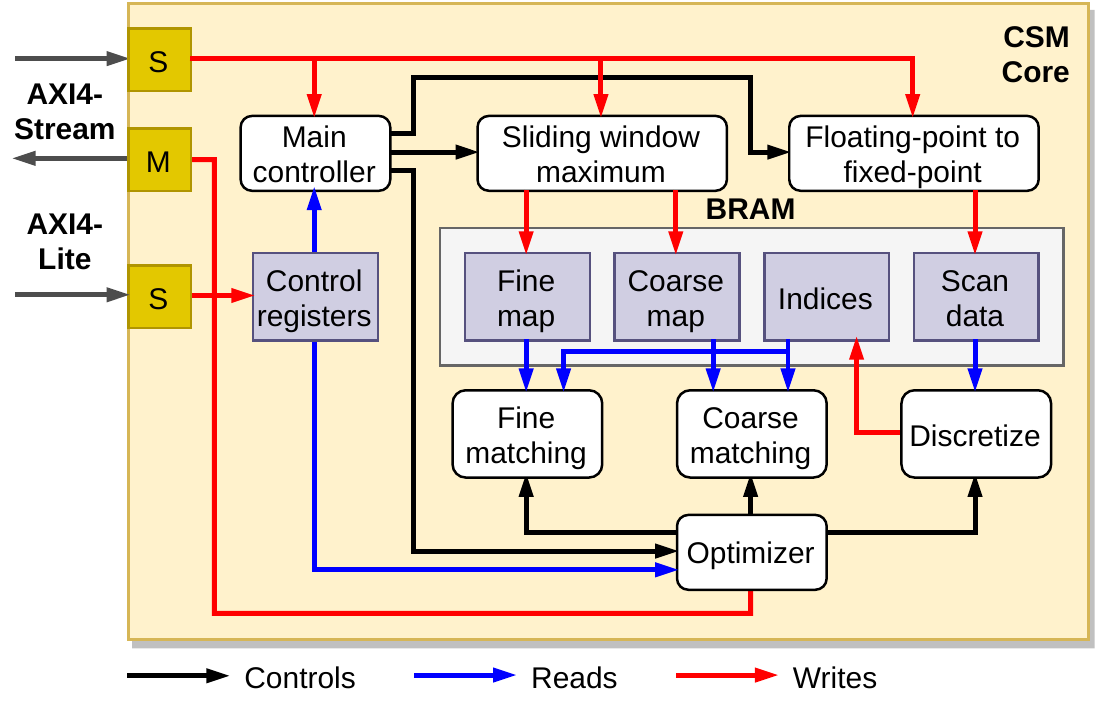}
    \caption{Block diagram of the CSM core}
    \label{fig:design-block-diagram-core}
\end{figure}

CSM core is comprised of a set of modules: (i) main controller, (ii) sliding window maximum, (iii) floating-point to fixed-point, (iv) optimizer, (v) scan discretization, (vi) coarse matching, and (vii) fine matching.
Figure \ref{fig:design-block-diagram-core} illustrates the data flow and interactions between these modules inside the core.
The functionality of each module is outlined as follows:
Module (i) reads the incoming data along with configurations and controls the flow of the CSM algorithm.

\begin{figure}[htbp]
    \centering
    \includegraphics[keepaspectratio,width=0.4\linewidth]{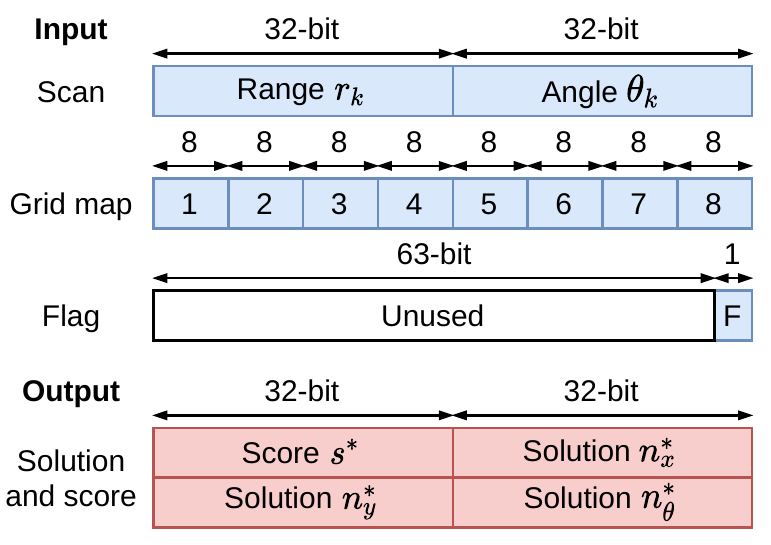}
    \caption{Input and output packet formats for data transfer between PS-PL}
    \label{fig:packet-format}
\end{figure}

Module (ii) retrieves the quantized grid map values $\mathcal{M}$ from the AXI4-Stream interface and stores them to the BRAM buffer (Fine map in Figure \ref{fig:design-block-diagram-core}).
It simultaneously applies a sliding window maximum filter (Equation \ref{eq:csm-sliding-window-maximum}) to the grid map values $\mathcal{M}$ to obtain a low-resolution grid map $\mathcal{M}'$, which is written to another BRAM buffer (Coarse map in Figure \ref{fig:design-block-diagram-core}).
The module retrieves map values in a row-wise order, stores column-wise maxima $\mathcal{M}''$ to a temporary cache (of size $320 \ \text{columns} \times w \ \text{rows}$), and then computes $\mathcal{M}'$ by taking row-wise maxima of values stored in the cache.
\begin{eqnarray}
    \mathcal{M}''(i, j) &=& \max_{j' \in \left[ 0, 1, \ldots, w - 1 \right]}
    \mathcal{M}(i, j + j') \\
    \mathcal{M}'(i, j) &=& \max_{i' \in \left[ 0, 1, \ldots, w - 1 \right]}
    \mathcal{M}''(i + i', j)
\end{eqnarray}
In our design, each 64-bit data packet coming from the streaming interface contains eight consecutive map values in 8-bit integer format as shown in Figure \ref{fig:packet-format}.
Module (iii) extracts two 32-bit floating-point values (range $r$ and angle $\theta$ of a scan point $\mathbf{z}$) from a 64-bit data packet (Figure \ref{fig:packet-format}), converts them to the 32-bit fixed-point values, and fills the scan buffer.
The scan buffer (Figure \ref{fig:design-block-diagram-core}) stores up to 512 range-angle pairs, i.e., $N \le 512$.

The optimizer module (iv) manages the coarse-to-fine search in CSM (Algorithm \ref{alg:csm}) using modules (v)-(vii).
For a given orientation $n_\theta$, the discretization module (v) computes the grid cell indices (Equation \ref{eq:scan-point-to-map}) corresponding to scan points and writes them back to the indices buffer.
The coarse (vi) and fine (vii) matching modules evaluate solution candidates (Equations \ref{eq:csm-evaluate-coarse}, \ref{eq:csm-evaluate-fine}) in parallel using the discretized scan indices and grid maps $\mathcal{M}, \mathcal{M}'$ located in the BRAM buffer.
After the matching is complete, the optimizer module (iv) returns back the optimal solution $n_x^*, n_y^*, n_\theta^*$ along with the score $s^*$ through the AXI4-Stream interface (Figure \ref{fig:packet-format}).
The following Sections \ref{sec:design-opt-buffer-size}-\ref{sec:design-opt-data-reuse} describe the design optimizations employed to exploit the inherent parallelism of CSM and realize the implementation on resource-constrained low-end FPGAs.

\subsection{Tuning the Grid Map Buffer Size} \label{sec:design-opt-buffer-size}
In the matching step (Algorithm \ref{alg:csm}, lines \ref{alg:csm-coarse-eval}, \ref{alg:csm-fine-eval}), the number of grid map accesses is $2 \widehat{w}_x \widehat{w}_y w_\theta \left( 1 + w^2 \right) N$ in the worst case, which reaches tens of millions in a typical setting.
In addition to this, as mentioned in Section \ref{sec:slam-csm}, score evaluations (Equations \ref{eq:csm-evaluate-fine}, \ref{eq:csm-evaluate-coarse}) exhibit irregular and unpredictable memory access patterns, which decrease the effectiveness of prefetching grid map values.
Reading only a necessary part of a grid map from DRAM and storing it to the small BRAM buffer will also degrade the performance due to the repetitive small data transfers from DRAM to BRAM.
This necessitates a BRAM storage for the entire map to minimize the memory access latency and the number of data transfers.

Storing the entire map on BRAM is, however, also infeasible, since it exhausts BRAM resources especially on small FPGA devices and increases a transfer overhead.
The grid map resolution needs to be fine enough for the precise pose estimation ($r = 0.05 \mathrm{m}$ in this paper), and contains tens of thousands of grid cells as a result.
For instance, grid maps in Figure \ref{fig:fine-and-coarse-map} contain 102,400 cells to cover $20.0 \times 12.8 \mathrm{m}$ area.

The proposed CSM core limits the size of grid maps up to $320 \times 320$ cells or $16 \times 16 \mathrm{m}$, which is sufficiently large considering the typical indoor environment (e.g., office).
From this limitation, the upper-bound of search space size becomes $16 \times 16 \mathrm{m}$ (i.e., $2 w_x, 2 w_y \le 320$ and $\widehat{w}_x, \widehat{w}_y \le 40$).
Only a region visible from a LiDAR is transferred to the core, which is reasonable considering the characteristics of rotating LiDAR sensors.
Since the measurement range is upper-bounded, only a fraction of the map around the current sensor position is necessary for matching: the other remaining part distant from the sensor can be omitted.
In the matching modules (vi)-(vii), scan points outside the boundary of trimmed grid maps are just ignored and not considered in the score evaluation.

Grid map values are also quantized to 6-bit (64 discrete values) due to the scarcity of BRAM slices: if 32-bit floating-point format is used, 65\% of BRAM available in Pynq-Z2 is consumed for each map buffer, and the above design (Figure \ref{fig:design-block-diagram-core}) becomes not synthesizable.
The CSM core stores only the high-order 6-bits of retrieved 8-bit values to the buffers.
As shown in the evaluation (Section \ref{sec:eval}), the algorithm performance is not severely affected by these optimizations and quantization errors.

\subsection{Parallelizing the Fine Matching} \label{sec:design-opt-fine-matching}
\begin{figure}[htbp]
    \centering
    \includegraphics[keepaspectratio,width=0.55\linewidth]{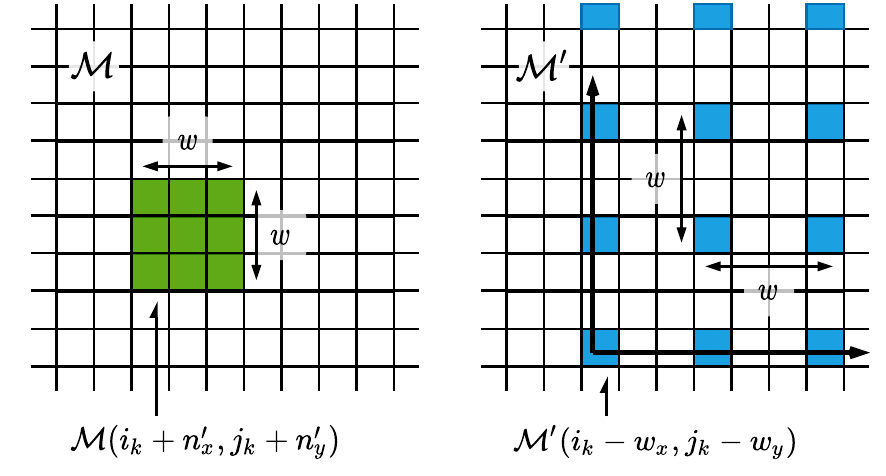}
    \caption{Access patterns in fine (left) and coarse (right) matching ($w = 3$)}
    \label{fig:matching-access-patterns}
\end{figure}

The matching modules (vi)-(vii) parallelize score evaluations by exploiting partially sequential access patterns found in CSM (Figure \ref{fig:matching-access-patterns}).
Fine matching (Algorithm \ref{alg:csm}, lines \ref{alg:csm-fine-loop-y}-\ref{alg:csm-solution-update}) evaluates matching scores for $w^2$ solution candidates ranging from $(n_x', n_y', n_\theta)$ to $(n_x' + w - 1, n_y' + w - 1, n_\theta)$ using Equation \ref{eq:csm-evaluate-fine}.
Observation of Equation \ref{eq:csm-evaluate-fine} reveals that, for the $k$-th scan point, grid map values from $\mathcal{M}(i_k + n_x', j_k + n_y')$ to $\mathcal{M}(i_k + n_x' + w - 1, j_k + n_y' + w - 1)$ are accessed (Figure \ref{fig:matching-access-patterns} (left)), to evaluate scores for the above $w^2$ candidates.

This motivates to develop the parallelized version of fine matching as shown in Algorithm \ref{alg:fine-matching-opt}, which firstly interchanges the loops over $k$ and $n_x$, and then completely unrolls the innermost $n_x$-loop by setting the unrolling factor to $w$, so that the spatial locality in map accesses is exploited.
It also parallelizes the loop over $n_y$.
By using Algorithm \ref{alg:fine-matching-opt}, our fine matching module can process $2w$ consecutive grid map elements and compute $2w$ scores $\left\{ s[\cdot, \cdot] \right\}$ in parallel, with only a small memory overhead.
With the cyclic partitioning of $\mathcal{M}$ along $x, y$ dimensions, the latency for fine matching is reduced from 235 to 15 $\mu \mathrm{s}$ ($w = 8, N = 360$), yielding the $15.74\times$ performance improvement.

\begin{algorithm}[htbp]
    \caption{Parallelized Fine Matching}
    \label{alg:fine-matching-opt}
    \begin{algorithmic}[1]
        \For{$n_y = n_y', n_y' + 2, \ldots, n_y' + w - 2$}
            \State $\forall i \in \left[ 0, \ldots, w - 1 \right], j \in \left[ 0, 1 \right], \ s[i, j] \gets 0$
            \For{$k = 1, \ldots, N$}
                \State $\forall i, j, \ s[i, j] \gets s[i, j] + \mathcal{M}(i_k + n_x' + i, j_k + n_y + j)$
            \EndFor
            \State $i^*, j^* \gets \argmax_{i, j} s[i, j]$
            \If{$s[i^*, j^*] > s^*$}
                \State $s^* \gets s[i^*, j^*]$
                \State $(n_x^*, n_y^*, n_\theta^*) \gets (n_x' + i^*, n_y + j^*, n_\theta)$
            \EndIf
        \EndFor
    \end{algorithmic}
\end{algorithm}

\subsection{Parallelizing the Coarse Matching} \label{sec:design-opt-coarse-matching}
\begin{figure}[htbp]
    \centering
    \includegraphics[keepaspectratio,width=0.55\linewidth]{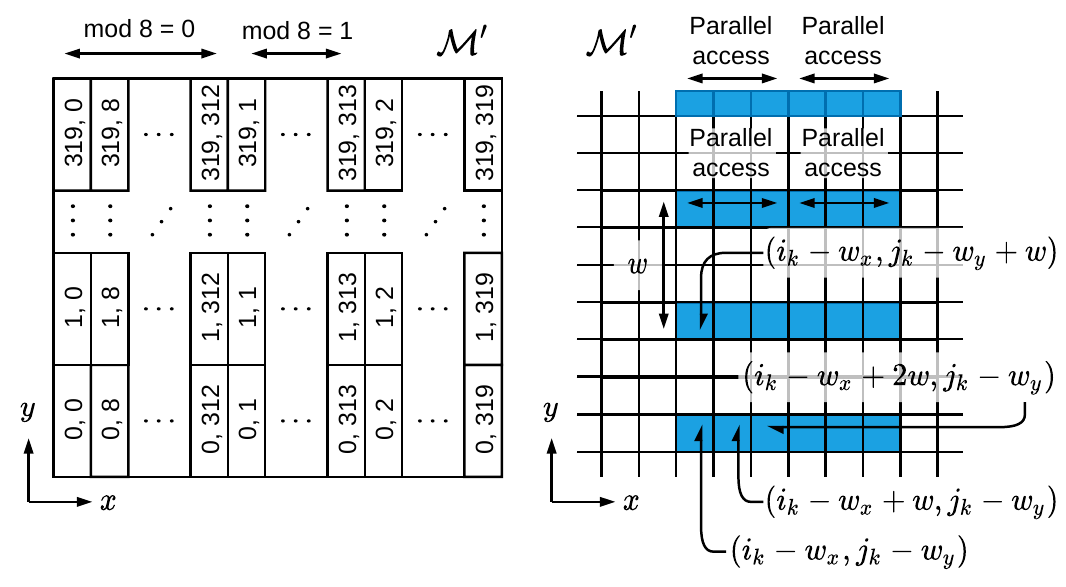}
    \caption{Rearranging the layout of coarse map $\mathcal{M}'$ (left, $w = 8$) to parallelize the strided accesses and optimized access patterns (right, $w = 3$)}
    \label{fig:coarse-map-layout}
\end{figure}

Our design also parallelizes coarse matching (Algorithm \ref{alg:csm}, lines \ref{alg:csm-coarse-loop-y}-\ref{alg:csm-coarse-eval-continue}), by reordering map elements and applying the similar optimizations as above.
For the $k$-th scan point, grid map elements starting from $\mathcal{M}(i_k - w_x, j_k - w_y)$ to $\mathcal{M}(i_k - w_x + (\widehat{w}_x - 1) w, j_k - w_y + (\widehat{w}_y - 1) w)$ are accessed when iterating over $n_x'$ and $n_y'$ (i.e., coarse search space), creating strided access patterns with the stride of $w$ (Figure \ref{fig:matching-access-patterns} (right)).
From this observation, parallelized coarse matching algorithm is obtained (Algorithm \ref{alg:coarse-matching-opt}), which unrolls the loop $n_x'$, thereby allowing the parallel evaluation of scores (Equation \ref{eq:csm-evaluate-coarse}).
Note that the horizontal order of coarse map elements is rearranged as depicted in Figure \ref{fig:coarse-map-layout} (left), to convert the stride accesses to non-stride sequential accesses (Figure \ref{fig:coarse-map-layout} (right)).
The unrolling factor of loop $n_x'$ is set to eight, meaning that eight consecutive elements along $x$-dimension are accessed and eight scores are computed in parallel (line \ref{alg:coarse-matching-opt-loop-i}).

\begin{algorithm}[htbp]
    \caption{Parallelized Coarse Matching}
    \label{alg:coarse-matching-opt}
    \begin{algorithmic}[1]
        \For{$n_y' = -w_y, -w_y + w, \ldots, -w_y + (\widehat{w}_y - 1) w$}
            \For{$n_x' = -w_x, -w_x + 8w, \ldots, -w_x + (\widehat{w}_x - 8) w$}
                \State $\forall i \in \left[ 0, \ldots, 7 \right], \ s'[i] \gets 0$
                \For{$k = 1, \ldots, N$}
                \label{alg:coarse-matching-opt-loop-k}
                    \State $\forall i, \ s'[i] \gets s'[i] + \mathcal{M}'(i_k - w_x + i \cdot w, j_k + n_y')$
                    \label{alg:coarse-matching-opt-loop-i}
                \EndFor
                \For{$i = 0, \ldots, 7$}
                    \If{$s'[i] > s^*$}
                        \State Perform fine matching (Algorithm \ref{alg:fine-matching-opt})
                    \EndIf
                \EndFor
            \EndFor
        \EndFor
    \end{algorithmic}
\end{algorithm}

\subsection{Reusing Previously Transferred Data} \label{sec:design-opt-data-reuse}
Using flag packets (Figure \ref{fig:packet-format}), the CSM core provides an option to skip data transfers, and reuses the previously transferred ones already stored on BRAM.
This improves the performance of one-to-many scan matching, i.e., matchings between a set of scans and a map $(\mathcal{M}, \left\{ \mathcal{S}_1, \ldots, \mathcal{S}_K \right\})$, or between a set of grid maps and a scan $(\left\{ \mathcal{M}_1, \ldots, \mathcal{M}_K \right\}, \mathcal{S})$.
In the first case, a map $\mathcal{M}$ is transferred only once before the matching with $\mathcal{S}_1$, and then is reused for matchings with $\mathcal{S}_2, \ldots, \mathcal{S}_K$, eliminating $K - 1$ unnecessary transfers of $\mathcal{M}$ and $K - 1$ precomputations of $\mathcal{M}'$ in the module (iii).
The next section describes the integration of CSM core into SLAM algorithms.

\begin{figure*}[htbp]
    \centering
    \includegraphics[keepaspectratio,width=\linewidth]{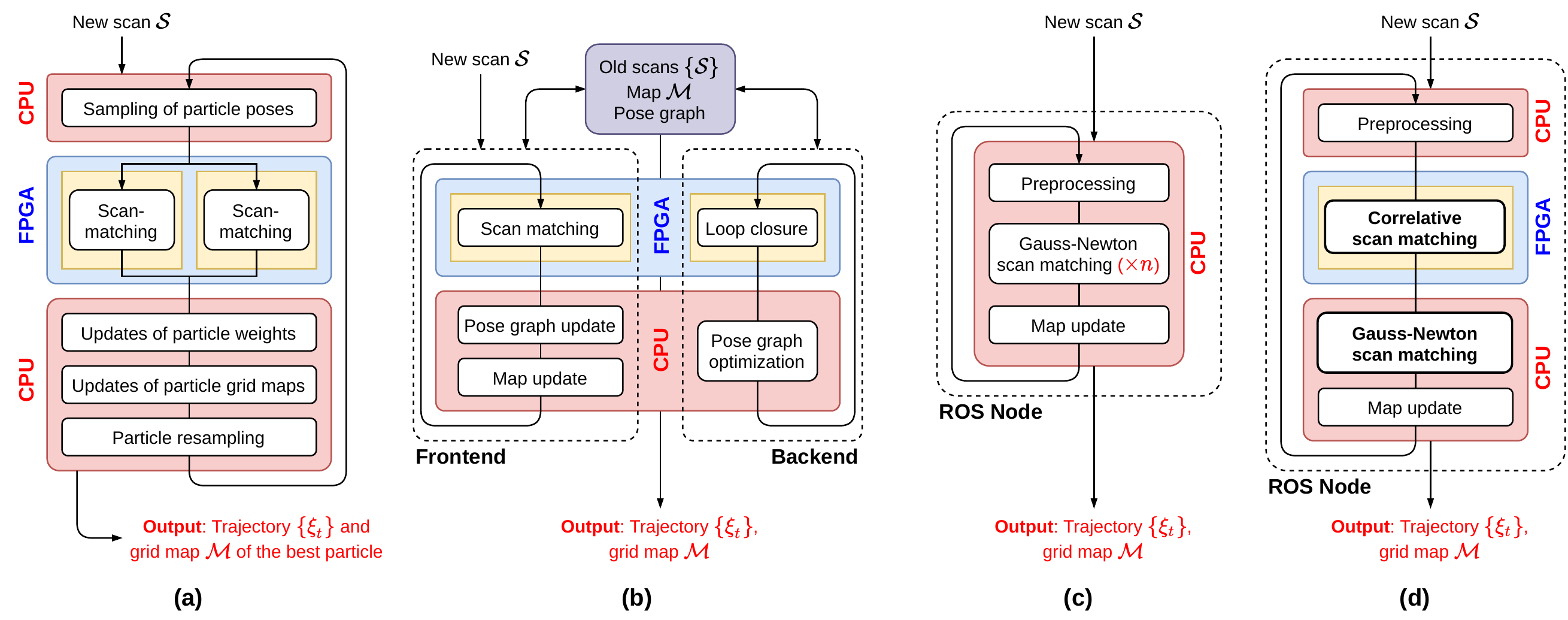}
    \vspace*{-15pt}
    \caption{(a) Particle filter-based SLAM (PF-SLAM) with FPGA acceleration, (b) graph-based SLAM with FPGA acceleration, (c) original version of Hector SLAM, (d) robustified version of Hector SLAM with FPGA acceleration.
    In PF-SLAM, we use two CSM cores to parallelize the scan matching for particles, where each CSM core handles scan matching for half of the particles.
    In graph-based SLAM, we employ one CSM core for the frontend scan matching, and the other core for the backend loop-closure detection.
    We robustify Hector SLAM by performing correlative scan matching (CSM) before iterative Gauss-Newton scan matching to avoid suboptimal solutions: CSM provides a pose estimate which is sufficiently close to the global optimum, and Gauss-Newton refines the estimate to obtain the final solution.
    We then utilize one CSM core to improve the performance of robustified Hector SLAM, which depends on ROS framework, while preserving the robustness and accuracy.
    In this way, we demonstrate that CSM core is ROS-compliant and supports the acceleration of ROS-based SLAM systems.}
    \label{fig:slam-system}
\end{figure*}

\subsection{Integration into PF- and Graph-based SLAM} \label{sec:design-application-to-slam}
CSM core is easily integrated into LiDAR SLAM systems that use scan-to-map matching.
As depicted in Figure \ref{fig:slam-system} (a), we offload the scan matching part in two algorithms, i.e., PF-SLAM and graph-based SLAM, using the CSM core interface.
In our PF-SLAM (Figure \ref{fig:slam-system} (a)) system, the interface creates two threads to parallelize scan matching for particles: each thread independently uses one pair of CSM core and DMA controller to process half of the particles, which involves the matching between one scan and multiple grid maps individually owned by particles.
The interface manages two CSM cores and makes use of flag packets (Figure \ref{fig:packet-format}) to avoid unnecessary transfers of the same scan.
Other tasks, e.g., updates of grid maps and particle resampling, are executed on the CPU side.

Our graph-based SLAM (Figure \ref{fig:slam-system} (b)) system is also multi-threaded and assigns one CSM core to each thread.
The system creates a set of submaps that together represent the entire grid map.
The frontend thread performs scan matching between the latest scan and a grid map consisting of recent scans to update the current robot pose.
The backend thread performs loop detections by attempting to match the recent scans against old submaps consisting of previously acquired scans, i.e., performs many-to-many scan matching.
In the algorithmic aspect, loop detection is basically the same as the frontend scan matching but with the larger search space, and thus the CSM core is also applicable to loop detections without modifying and re-synthesizing the programmable logic part.
The backend treats a matching attempt as a success, if a returned score $s^*$ is greater than the predefined threshold $s_T$.
We implement a CSM core interface for graph-based SLAM, which manages one CSM core and works for both the frontend scan matching and the backend loop detection.

\subsection{Robustified Hector SLAM} \label{sec:design-robustified-hector-slam}
Hector SLAM~\cite{Kohlbrecher11} is another famous method, and is classified as a scan matching-based SLAM.
Algorithm \ref{alg:original-hector-slam} presents the main steps of Hector SLAM.
It aligns an input LiDAR scan with an entire occupancy grid map to estimate the ego-motion of the robot, and incrementally updates the current robot pose.
This so-called dead reckoning is prone to error accumulation and noises in sensor measurements, as it adopts a scan-to-map matching method derived from Gauss-Newton, which has inherent weaknesses as discussed in Section \ref{sec:slam-csm}.
To mitigate this issue, Hector SLAM keeps multiple grid maps with different resolution levels.
Specifically, it creates a pyramid of $n$ grid maps $\mathcal{M}_0, \mathcal{M}_1, \ldots, \mathcal{M}_{n - 1}$ with exponentially decreasing resolutions of $r, 2r, 4r, \ldots, 2^{n - 1} r$.
The original scan matcher then follows a coarse-to-fine alignment strategy (lines \ref{alg:original-hector-slam-gaussnewton-begin}-\ref{alg:original-hector-slam-gaussnewton-end}), where the pose estimate in the previous stage is fed into the next finer stage as an initial guess.
This multi-resolution map approach comes at the cost of increased memory consumption and computational cost for storing and updating multiple grid maps (line \ref{alg:original-hector-slam-map}).
In addition, Hector SLAM is still not able to achieve satisfactory accuracy, as shown in the evaluation (Table \ref{tbl:comparison-trajectory-hectorslam}).

We therefore propose to combine CSM and Gauss-Newton (Figure \ref{fig:slam-system} (d)) instead of only using Gauss-Newton (Figure \ref{fig:slam-system} (c)).
Algorithm \ref{alg:robustified-hector-slam} summarizes the modified version of Hector SLAM.
Firstly, we employ CSM (line \ref{alg:robustified-hector-slam-csm}) to find an optimal pose $\bm{\xi}_t'$ at $r = 5 \mathrm{cm}$ and $\delta_\theta \ \mathrm{rad}$ accuracy (i.e., search step size) inside a discretized search space, which is centered at an initial guess, e.g., a pose obtained from the last scan matching execution $\bm{\xi}_{t - 1}$.
Then, we use Gauss-Newton (line \ref{alg:robustified-hector-slam-gaussnewton}) to obtain the final pose estimate $\bm{\xi}_t$ in the continuous search space.
We can expect that it converges to a globally-optimal solution $\bm{\xi}^*$, since search window strategy in CSM helps to avoid suboptimal solutions, and consequently the intermediate result $\bm{\xi}_t'$ is sufficiently close to $\bm{\xi}^*$.
In this way, Gauss-Newton acts as a refinement step to the CSM.
Since this two-step approach is robust, we do not apply the multi-resolution map approach as explained above, i.e., we only store the finest grid map $\mathcal{M}$ (with a resolution of $r$) and perform scan matching only once per LiDAR scan.
This robustified version of Hector SLAM (Figure \ref{fig:slam-system} (d)) offers better accuracy than the original one as shown in Table \ref{tbl:comparison-trajectory-hectorslam}, while at the same time saving memory space and computational cost for keeping grid maps.

\begin{algorithm}[h]
  \caption{Original Hector SLAM}
  \label{alg:original-hector-slam}
  \begin{algorithmic}[1]
    \State Create grid maps $\mathcal{M}_0, \mathcal{M}_1, \ldots, \mathcal{M}_{n - 1}$ with exponentially decreasing resolutions $r, 2r, \ldots, 2^{n - 1}r$
    \State Initialize a current pose: $\bm{\xi}_0 \gets \left[ 0, 0, 0 \right]$
    \For{$t = 1, \ldots, $}
      \State Receive a new LiDAR scan $\mathcal{S}_t$ and remove outliers
      \LeftComment{\textbf{Gauss-Newton scan matching}}
      \State $\bm{\xi}_{t, n - 1} \gets \bm{\xi}_{t - 1}$ \label{alg:original-hector-slam-gaussnewton-begin}
      \For{$i = n - 1, n - 2, \ldots, 0$}
        \State \parbox[t]{.9\linewidth}{
          Perform Gauss-Newton scan matching using $\mathcal{M}_i$.
          Improve pose estimate $\bm{\xi}_{t, i}$ according to the cost function $J(\bm{\xi}_{t, i})$ defined as follows: \\
          \hspace*{\algorithmicindent}
          $J(\bm{\xi}_{t, i}; \mathcal{M}_i, \mathcal{S}) = \displaystyle \sum_{k = 1}^N
          \left( 1 - \mathcal{M}_i(h(\bm{\xi}_{t, i}, \bm{z}_k)) \right)^2$ \\
          Resulting pose estimate is used as the initial estimate for the next iteration ($\bm{\xi}_{t, i - 1} \gets \bm{\xi}_{t, i}$).}
      \EndFor \label{alg:original-hector-slam-gaussnewton-end}
      \State Update the current pose: $\bm{\xi}_t \gets \bm{\xi}_{t, 0}$
      \LeftComment{\textbf{Map update}}
      \For{$i = 0, \ldots, n - 1$}
        \State Update $\mathcal{M}_i$ using scan $\mathcal{S}_t$ and pose $\bm{\xi}_t$ \label{alg:original-hector-slam-map}
      \EndFor
    \EndFor
    \State \Return Trajectory $\left\{ \bm{\xi}_0, \bm{\xi}_1, \ldots \right\}$, grid map $\mathcal{M}_0$
  \end{algorithmic}
\end{algorithm}

\begin{algorithm}[h]
  \caption{Robustified Hector SLAM}
  \label{alg:robustified-hector-slam}
  \begin{algorithmic}[1]
    \State Create a grid map $\mathcal{M}$ with a resolution of $r$
    \State Initialize a current pose: $\bm{\xi}_0 \gets \left[ 0, 0, 0 \right]$
    \For{$t = 1, \ldots, $}
      \State Receive a new LiDAR scan $\mathcal{S}_t$ and remove outliers
      \State \parbox[t]{.9\linewidth}{
        Perform CSM with the initial guess $\bm{\xi}_{t - 1}$ and obtain a pose estimate $\bm{\xi}_t'$} \label{alg:robustified-hector-slam-csm}
      \State \parbox[t]{.9\linewidth}{
        Refine the pose estimate $\bm{\xi}_t'$ using Gauss-Newton and obtain a final pose estimate $\bm{\xi}_t$. The cost function $J(\bm{\xi}_t')$ is defined as follows: \\
        \hspace*{\algorithmicindent}
        $J(\bm{\xi}_t'; \mathcal{M}, \mathcal{S}) = \displaystyle \sum_{k = 1}^N
        \left( 1 - \mathcal{M}(h(\bm{\xi}_t', \bm{z}_k)) \right)^2$} \label{alg:robustified-hector-slam-gaussnewton}
      \State Update $\mathcal{M}$ using scan $\mathcal{S}_t$ and pose $\bm{\xi}_t$
    \EndFor
    \State \Return Trajectory $\left\{ \bm{\xi}_0, \bm{\xi}_1, \ldots \right\}$, grid map $\mathcal{M}$
  \end{algorithmic}
\end{algorithm}

\subsection{Integration into Robustified Hector SLAM} \label{sec:design-application-to-hector-slam}
In this section, we extend the proposed FPGA-based heterogeneous platform to support ROS (Robot Operating System) applications for improved practicality and usability.
ROS middleware provides a variety of frameworks and tools to ease the development of robotic applications.
It is often a basic building block for various LiDAR SLAM systems, e.g., GMapping~\cite{ROSSLAMGMapping}, Google Cartographer~\cite{Hess16}, and Hector SLAM~\cite{ROSHectorSLAM}.

We introduce a ROS-based interface for the robustified Hector SLAM (Section \ref{sec:design-robustified-hector-slam}) to handle the interactions between CSM cores and ROS nodes.
This ROS-FPGA interface retrieves a set of parameters, e.g., physical start address and length of the memory-mapped regions, from a ROS parameter server, so that ROS nodes can directly change the behavior of the interface.
It provides a function to perform preliminary tasks from ROS nodes to set up CSM cores: the function allocates contiguous memory for DMA transfers, maps control registers into physical address spaces, and initializes CSM cores together with DMA controllers through the registers.
In response to a request from Hector SLAM, the interface employs one CSM core and returns a pose estimate $\bm{\xi}'$, which is then passed on to the Gauss-Newton scan matcher as a good starting point to compute the final pose estimate $\bm{\xi}^*$, as described above.
Note that we could use two CSM cores to further improve the efficiency: given a search window, each core tries to find an optimal solution from the half of the search window.
After the optimal poses and scores $(\bm{\xi}_1^*, s_1^*), (\bm{\xi}_2^*, s_2^*)$ are returned from two CSM cores, the pose associated with the larger score is returned as the final pose estimate $\bm{\xi}^*$.
The next section evaluates the performance of CSM core using three SLAM implementations and real-world datasets.

\section{Implementation and Experimental Results} \label{sec:eval}
\subsection{Details of the CSM Core Implementation} \label{sec:eval-impl-core}
We developed the proposed scan matching core in C++ using Xilinx Vivado HLS 2019.2, and run synthesis and place-and-route using Vivado 2019.2.
Pynq-Z2 development board (Figure \ref{fig:board-and-robot} (left)) is chosen as a target device, to show that our design fits within the low-priced and resource-constrained FPGAs.
Pynq-Z2 consists of Xilinx XC7Z020-1CLG400C FPGA fabric (equivalent to Artix-7), dual-core ARM Cortex-A9 CPU at 650MHz, and 512MB DDR3 DRAM.
It runs Pynq Linux (version 2.5) based on Ubuntu 18.04.
The operation frequency of the CSM core is set to 100 MHz.

\subsection{Details of the SLAM Implementations} \label{sec:eval-impl-slam}
Our PF-SLAM and graph-SLAM systems were implemented in C++ from scratch without ROS (Robot Operating System).
We compiled them using GCC 7.3.0 with \texttt{-O3} compiler flag for full optimization.
Our PF-SLAM system utilizes OpenMP to take advantage of the dual-core CPU and parallelize the software scan matching.
We used 16 particles throughout the experiments presented below.
The graph-SLAM backend uses g$^2$o library~\cite{Kuemmerle11} for pose graph optimization.

As mentioned in Section \ref{sec:slam-csm}, CSM estimates the robot position at $r = 5 \mathrm{cm}$ accuracy, which is same as the grid map resolution.
After performing CSM, our SLAM system refines the pose estimate $\bm{\xi}^*$ at a subpixel accuracy using iterative scan matching methods.
Our PF-SLAM and graph-based SLAM systems adopt hill-climbing and gradient-based methods for the additional pose refinement, respectively.

As described in Section \ref{sec:design-robustified-hector-slam}, we modified the source code of Hector SLAM, which is distributed as an open-source ROS package~\cite{ROSHectorSLAM}, to robustify the scan matching part.
Specifically, we used the source code of the latest version (0.5.2) and added an option to use CSM instead of Gauss-Newton.
The pose estimate returned from CSM is used as a good starting point for Gauss-Newton scan matcher.
Note that we also set the number of grid maps to one when using CSM, since we do not take the multi-resolution approach.
We then integrated the proposed CSM cores to the robustified Hector SLAM using the ROS-FPGA interface introduced in Section \ref{sec:design-application-to-hector-slam}.

We run the original and robustified version of Hector SLAM on ROS Melodic, which is compatible with Pynq Linux.
In the evaluation, a set of ROS nodes are launched as separate processes, and they communicate via a publisher-subscriber message passing: Hector SLAM itself is implemented as a single ROS node (Figure \ref{fig:slam-system}), subscribes to LiDAR scan messages from another node, and publishes odometry and grid map messages to the other node.
We excluded the overhead for communication between nodes, and only measured a wall-clock execution time inside the Hector SLAM node for a fair performance comparison.

Pynq Linux provides a Python library to interact with peripherals (e.g., SPI, I2C) and custom IP modules on the PL part.
Since our SLAM systems and Hector SLAM are written in C++, we ported a necessary part of the library to C++ for configuring the PL part and communicating with CSM cores and DMA controllers.
We used FPGA Manager and CMA (Contiguous Memory Allocator) in Linux kernel to load bitstreams and allocate contiguous memory spaces, respectively.


\subsection{Real-world Datasets}
\begin{figure}[htbp]
    \centering
    \includegraphics[keepaspectratio,width=0.55\linewidth]{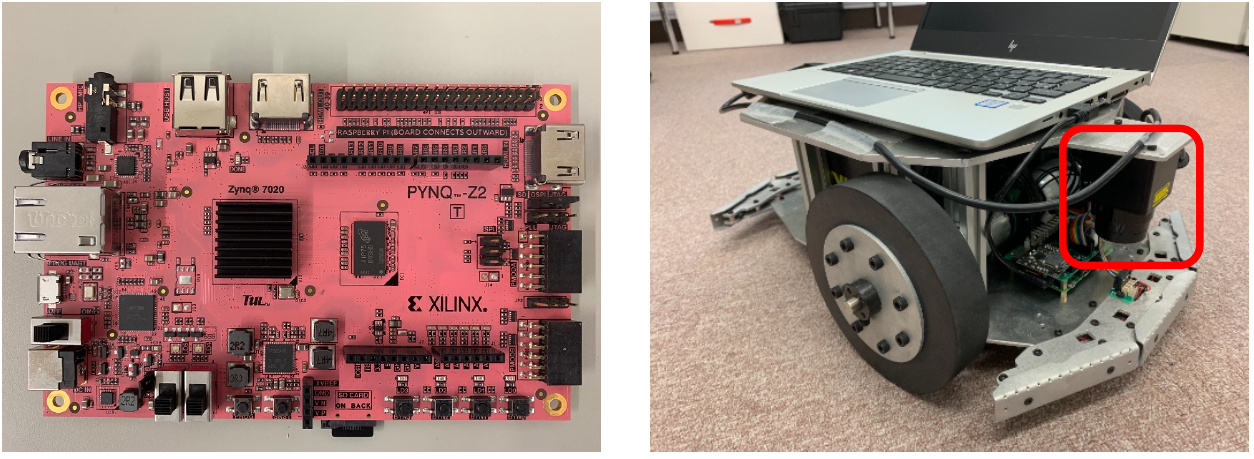}
    \caption{Pynq-Z2 development board (left) and wheeled mobile robot equipped with Hokuyo URG-04LX-UG01 LiDAR sensor (red square) (right)}
    \label{fig:board-and-robot}
\end{figure}

As real-world datasets, we used three publicly available datasets from the Radish repository~\cite{RadishDataset}, recorded at Intel Research Lab (\textbf{Intel}, $28.5 \times 28.5 \mathrm{m}$), MIT CSAIL Building (\textbf{MIT}, $61 \times 46.5 \mathrm{m}$), and ACES Building (\textbf{ACES}, $56 \times 58 \mathrm{m}$).
We also obtained a dataset at the corridor in our university campus building using a Hokuyo URG-04LX-UG01 range finder, which has a maximum measurement range of $5.6 \mathrm{m}$ and an angular resolution of $0.36^\circ$ (Figure \ref{fig:board-and-robot} (right)).
The wheeled robot was controlled remotely to move around the corridor twice for 1,226 seconds, and collected scans and wheel odometry data at around 500ms interval (2Hz).
Additionally, we used a \textbf{Revo LDS} dataset from \cite{Hess16}, which contains a set of LiDAR scans captured at around 200ms interval (5Hz) using the low-cost Revo LDS LiDAR.


\subsection{Execution Time Breakdown and Latency} \label{sec:eval-latency}
\begin{figure}[htbp]
    \centering
    \includegraphics[keepaspectratio,width=0.55\linewidth]{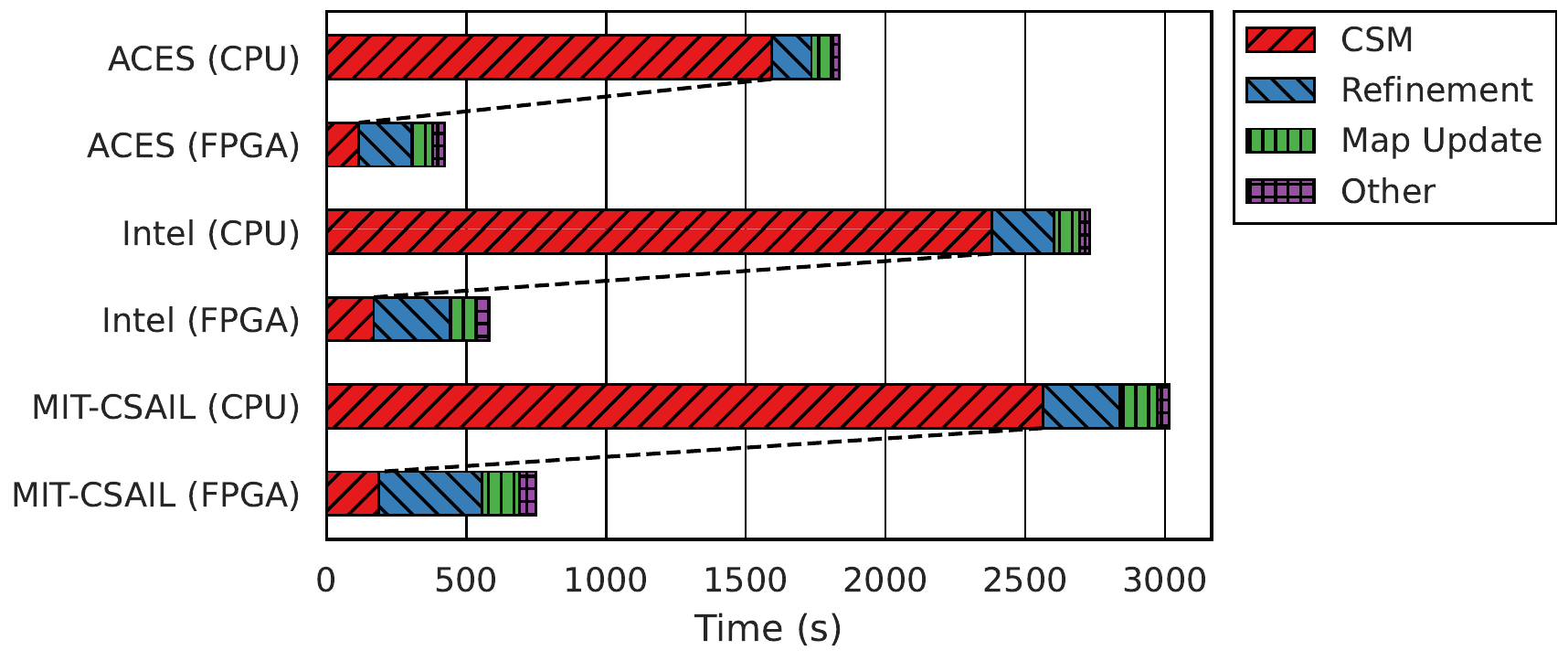}
    \caption{Execution time breakdown of PF-SLAM.
    In \textbf{Intel} dataset, our PF-SLAM implementation with FPGA acceleration (4th row) took 584.4s (416.2ms per frame on average) to complete, which is 4.67x speedup compared to the CPU-only version (3rd row).
    \textbf{Refinement} refers to the time taken to refine pose estimates returned from CSM using gradient-based scan matching, while \textbf{Map Update} is the total time taken to update grid maps of all particles.
    We set the search window size of CSM to (0.25m, 0.25m, 0.25rad) to obtain the best accuracy for these datasets.}
    \label{fig:gmapping-speedup}
\end{figure}

\begin{figure}[htbp]
    \centering
    \includegraphics[keepaspectratio,width=0.55\linewidth]{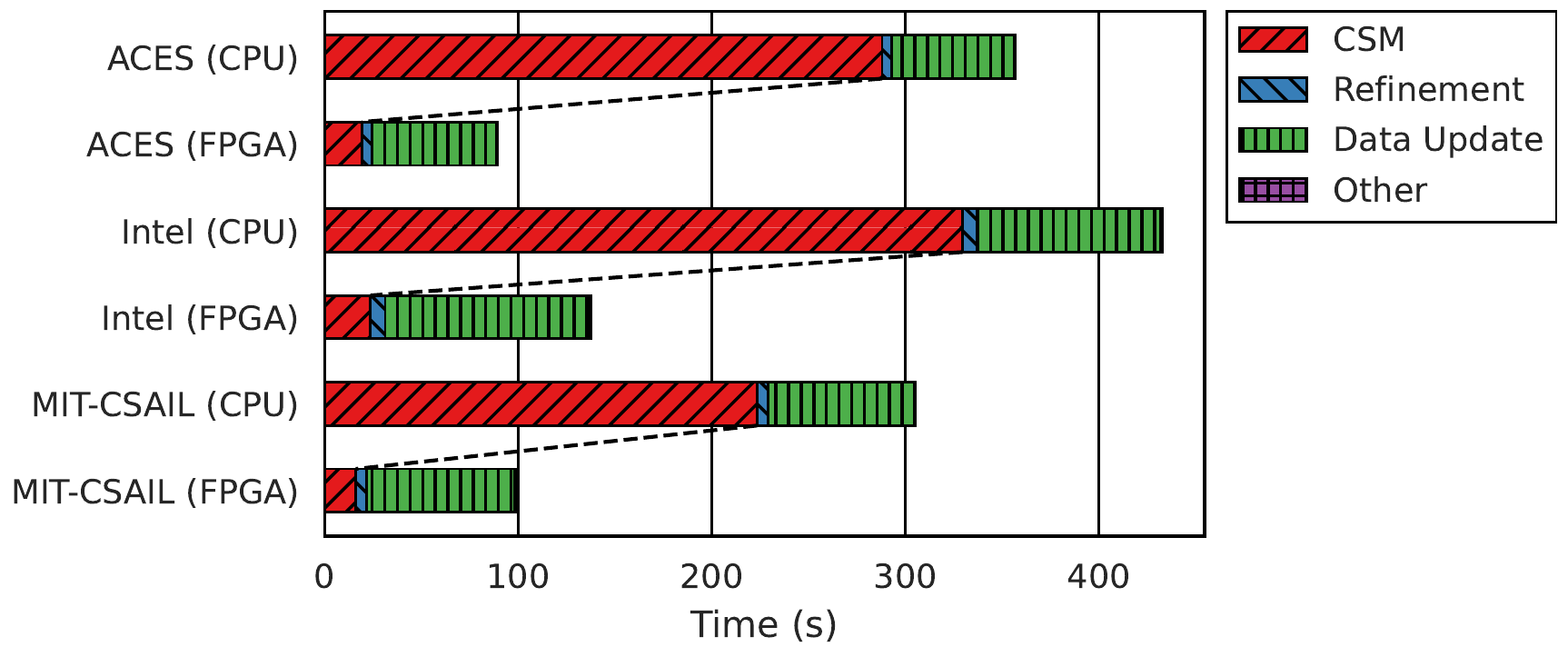}
    \caption{Execution time breakdown of graph-based SLAM frontend.
    Our graph-based SLAM implementation took 137.9s (98.2ms per frame on average) to process \textbf{Intel} dataset when using our CSM cores (4th row), achieving 3.14x speedup compared to the CPU-only version (3rd row).
    \textbf{Data Update} refers to the time it took for SLAM frontend to update grid map and pose graph for the next LiDAR scan, and to synchronize data with the SLAM backend.
    We chose the search window size of CSM to be (0.25m, 0.25m, 0.5rad) to obtain the best accuracy for these datasets.}
    \label{fig:graphslam-frontend-speedup}
\end{figure}

\begin{figure}[htbp]
    \centering
    \includegraphics[keepaspectratio,width=0.55\linewidth]{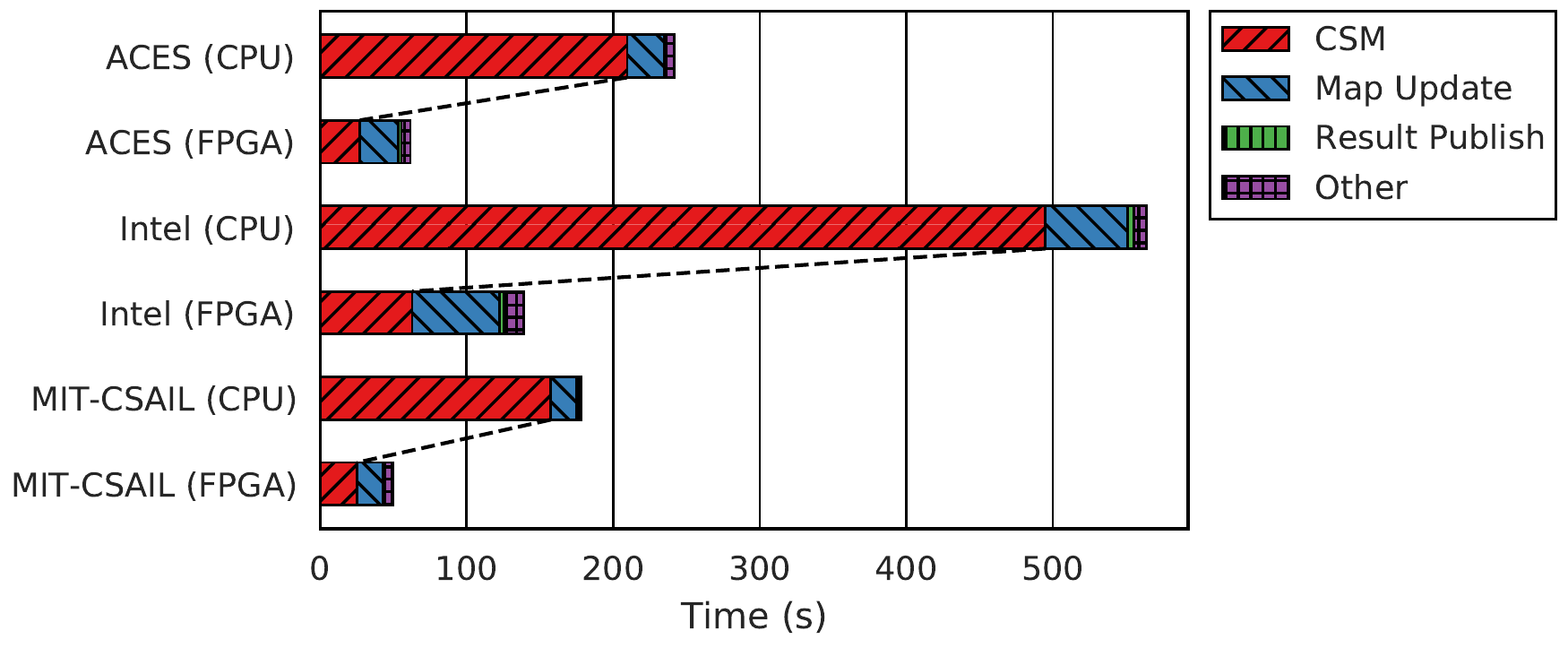}
    \caption{Execution time breakdown of Hector SLAM.
    In \textbf{Intel} dataset, the running time of Hector SLAM with our FPGA accelerator (4th row) was 139.0s (82.4ms per frame on average), which is 4.06x faster than the CPU-only version (3rd row).
    \textbf{Result Publish} is the time required for publishing resultant pose estimates and coordinate transformations as ROS messages.
    The search window size of CSM is set to (0.25m, 0.25m, 0.25rad) for \textbf{ACES} and \textbf{MIT-CSAIL}, and (0.25m, 0.25m, 0.5rad) for \textbf{Intel} to achieve the best accuracy.
    }
    \label{fig:hectorslam-speedup}
\end{figure}

Figures \ref{fig:gmapping-speedup}-\ref{fig:hectorslam-speedup} show execution time breakdowns of PF-SLAM, graph-based SLAM frontend, and Hector SLAM when using software or FPGA-based CSM implementation.
These evaluations consider the data transfer overhead between PS-PL for fair comparisons.
In PF-SLAM, CSM core speeds up the scan matching process by 13.67x, 14.09x, and 13.67x, reducing the total execution time by 4.34x, 4.67x, and 4.03x on \textbf{ACES}, \textbf{Intel}, and \textbf{MIT-CSAIL}, respectively.
Similarly, in graph-based SLAM, CSM core speeds up the frontend scan matching by 14.84x, 13.90x, and 13.85x on those datasets, thus achieving 4.00x, 3.14x, and 3.09x reduction in total execution time.
In Hector SLAM, scan matching part is accelerated by 7.72x, 7.88x, and 6.21x on those datasets, which contributes to the overall speedup of 3.92x, 4.06x, and 3.57x.
The results above indicate the effectiveness of CSM core on a variety of datasets and a wide range of SLAM methods.

The performance improvement is achieved by parallelizing the coarse-to-fine matching as described in Section \ref{sec:design}, and by offloading coarse map precomputations to the CSM core.
The computational complexity of this precomputation is linear with the number of grid cells in a given map, and has the measurable impact on the entire performance: in PF-SLAM and graph-based SLAM, this accounted up to 36.9\% and 45.1\% of the entire scan matching process.

Compared to PF-SLAM that maintains a set of particles and considers multiple hypotheses on robot trajectory and map, graph-based SLAM computes only the most plausible estimate, leading to better wall-clock times.
Hector SLAM has the minimum wall-clock time for each iteration owing to its algorithmic simplicity.
It solely performs scan matching for every incoming LiDAR scan, whereas graph-based SLAM frontend builds a pose graph as well as a map, and also synchronizes these data with SLAM backend in order to receive an optimized pose graph and perform loop-closure detection with new LiDAR scans.
Figures \ref{fig:gmapping-speedup}-\ref{fig:hectorslam-speedup} highlight the performance advantage of Hector SLAM and graph-based SLAM over PF-SLAM (e.g., 82.4ms, 98.2ms, and 416.2ms per frame in \textbf{Intel}).

\begin{figure}[htbp]
    \centering
    \includegraphics[keepaspectratio,width=0.55\linewidth]{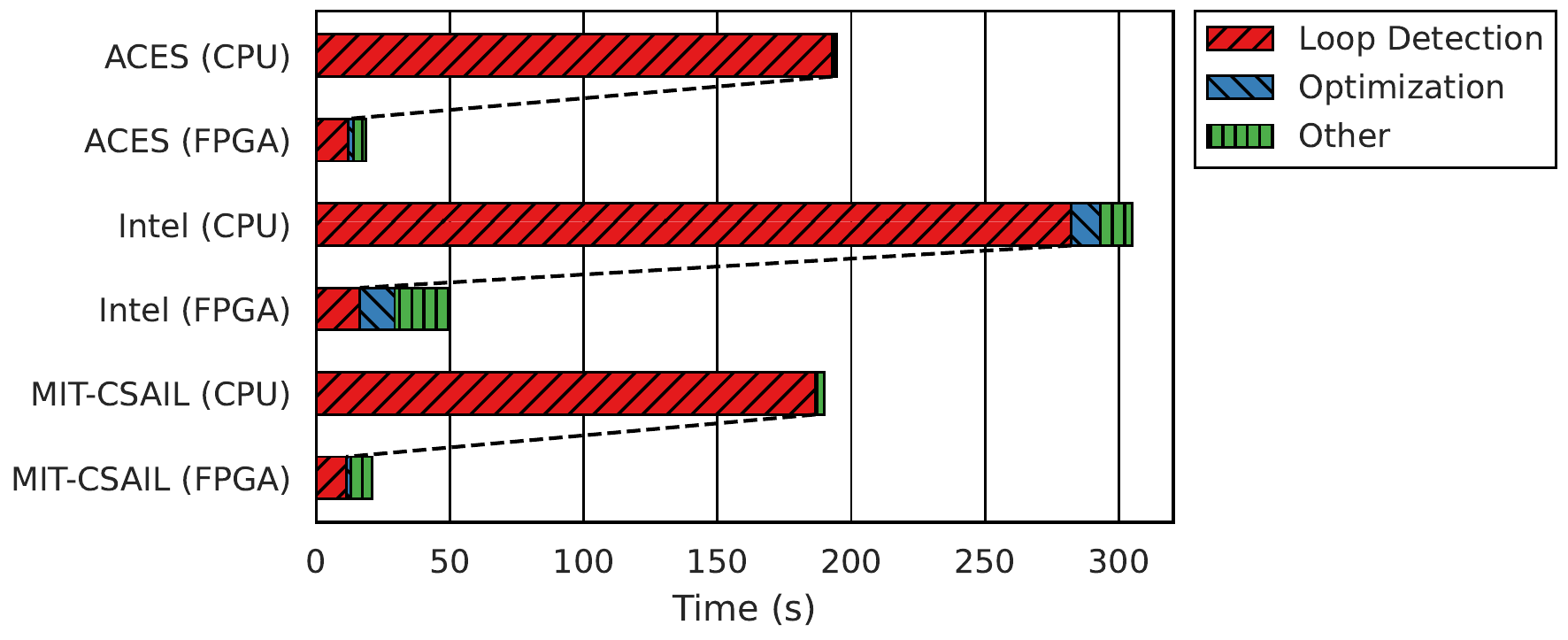}
    \caption{Execution time breakdown of graph-based SLAM backend.
    \textbf{Optimization} refers to the pose graph optimization which is performed after successful loop detections to refine the whole trajectory.
    The loop detection part is sped up by 16.18x, 17.23x, and 16.64x, resulting in the reduction of total processing time by 10.41x, 6.10x, and 9.10x for these datasets.}
    \label{fig:graphslam-backend-speedup}
\end{figure}

Figure \ref{fig:graphslam-backend-speedup} compares the total execution time with and without FPGA acceleration.
In graph-based SLAM backend, the execution time for all successful loop detections is reduced by 18.92x, 16.14x, and 6.65x on \textbf{ACES}, \textbf{Intel}, and \textbf{MIT-CSAIL} datasets, respectively.
CSM core offered better loop detection capability than the software counterpart: it detected 111.8, 475.0, and 111.7 loops on average, whereas the software version detected 42.0, 399.4, and 47.4 loops.
Putting these results together, loop detection is accelerated by 50.34x (3765.0ms to 66.55ms per successful detection on average), 19.19x (480.62ms to 21.34ms), and 15.69x (523.6ms to 28.32ms), respectively.
The search window size along $x, y, \theta$ axes is set to (5.0m, 5.0m, 0.75rad) for \textbf{ACES}, and (2.5m, 2.5m, 0.5rad) for the other two, thus indicating that CSM core achieves better performance improvements when using a larger search window.
Note that the comparisons presented here are only indicative, since different grid maps and LiDAR scans are used for loop detections on CPU and FPGA due to the timing differences.

The three datasets (\textbf{ACES}, \textbf{Intel}, \textbf{MIT-CSAIL}) used contain 795, 1404, and 686 LiDAR scans; SLAM needs to process all scan inputs in less than 147.2s, 277.2s, and 146.3s, i.e., data acquisition time, for real-time performance.
As seen in Figures \ref{fig:graphslam-frontend-speedup} and \ref{fig:hectorslam-speedup}, the proposed CSM core enable real-time processing speed in both graph-based SLAM and Hector SLAM.
When using \textbf{Revo LDS} dataset, the latency for processing a single scan was $63.7 \pm 39.4 \mathrm{ms}$ in graph-based SLAM, which is shorter than the scan period (i.e., 200ms), meaning that the real-time performance is achieved for most of the runtime.

\begin{figure*}[h!]
    \centering
    \includegraphics[keepaspectratio,width=\linewidth]{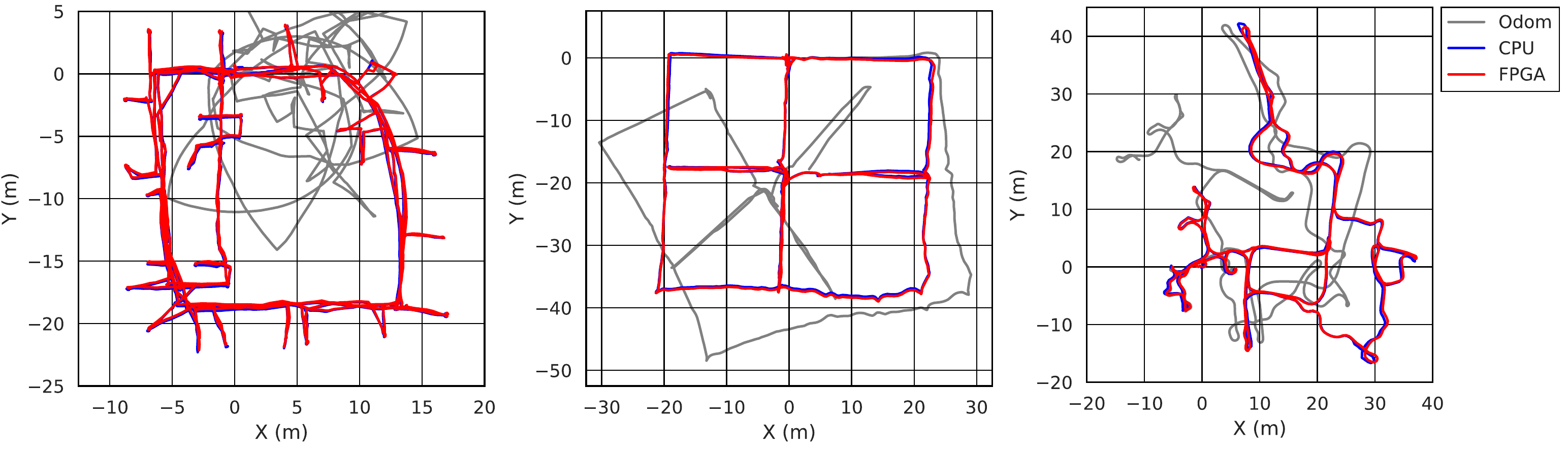}
    \caption{Comparison of the trajectories (left: PF-SLAM + \textbf{Intel}, center: Graph-based SLAM + \textbf{ACES}, right: Hector SLAM + \textbf{MIT-CSAIL})}
    \label{fig:trajectory-comparisons}
\end{figure*}

\begin{figure}[htbp]
    \centering
    \includegraphics[keepaspectratio,width=0.55\linewidth]{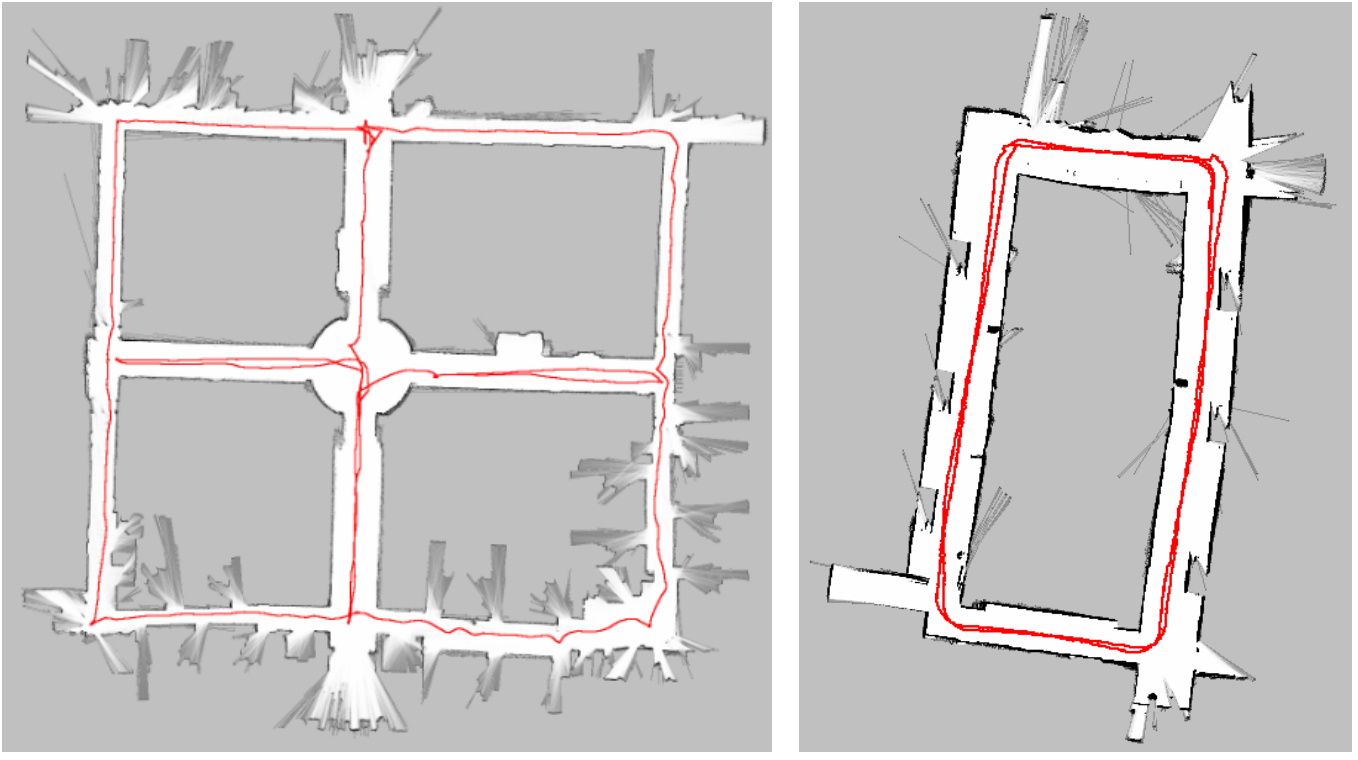}
    \caption{Grid map and robot trajectory (red line) obtained using our proposed CSM core (left: graph-based SLAM and \textbf{ACES} dataset, right: graph-based SLAM and dataset captured at the corridor)}
    \label{fig:grid-map-aces-sousoukan}
\end{figure}

\subsection{Accuracy} \label{sec:eval-accuracy}
For qualitative evaluation, Figures \ref{fig:trajectory-comparisons} and \ref{fig:grid-map-aces-sousoukan} show the grid maps obtained from CSM cores.
Note that the grid map for \textbf{Revo LDS} dataset is created using only LiDAR scans without odometry information.
Figure \ref{fig:trajectory-comparisons} compares the robot trajectories obtained from CSM cores against the ones from software scan matchers and pure odometry (i.e., only using wheel encoders to estimate the ego-motion of the robot, gray line).
We can observe the significant overlap between two trajectories (CPU and FPGA, blue and red lines), which indicates that the output of CSM core preserves the accuracy of the software scan matchers, in spite of the quantization of grid map values, limitation of the buffer sizes, and rounding errors introduced by fixed-point arithmetic operations.
Figure \ref{fig:grid-map-aces-sousoukan} (right) shows the grid map of the corridor: we can confirm that graph-based SLAM successfully detected and closed a loop when a robot moved around the corridor.

For quantitative comparison, we present the accuracy of robot trajectories obtained from Radish datasets in Table \ref{tbl:comparison-trajectory}.
Since ground-truth trajectories are not available in Radish datasets, we used the accuracy metric introduced in \cite{Kuemmerle09} that does not require them.
This metric is interpreted as the relative pose error (RPE) between an output robot trajectory and a handcrafted ground-truth trajectory.

The metric is defined as translational and rotational errors $\varepsilon_\mathrm{trans}, \varepsilon_\mathrm{rot}$ in robot poses along a given trajectory:
\begin{eqnarray}
    \varepsilon_\mathrm{trans}^n &=& \frac{1}{\left| \mathcal{A} \right|}
    \sum_{(i, j) \in \mathcal{A}} \mathrm{trans}(\bm{\Delta \xi}_{i, j} \ominus \bm{\Delta \xi}_{i, j}^*)^n \label{eq:accuracy-metric-trans} \\
    \varepsilon_\mathrm{rot}^n &=& \frac{1}{\left| \mathcal{A} \right|}
    \sum_{(i, j) \in \mathcal{A}} \mathrm{rot}(\bm{\Delta \xi}_{i, j} \ominus \bm{\Delta \xi}_{i, j}^*)^n, \label{eq:accuracy-metric-rot}
\end{eqnarray}
where $n = 1, 2$, and $\mathcal{A}$ is a set of relative relations.
$\bm{\Delta \xi}_{i, j} = \bm{\xi}_j \ominus \bm{\xi}_i$ denotes a robot movement (i.e., relative pose) from time $i$ to $j$, obtained from the output trajectory $\left\{ \bm{\xi}_0, \bm{\xi}_1, \ldots, \bm{\xi}_T \right\}$.
$\bm{\Delta \xi}_{i, j}^*$ is the ground-truth robot motion, which is obtained by manually superimposing LiDAR scans $\mathcal{S}_i, \mathcal{S}_j$ at time $i$ and $j$ using the dedicated software.
$\mathrm{trans}(\bm{\Delta \xi})$ and $\mathrm{rot}(\bm{\Delta \xi})$ compute a distance $\sqrt{\Delta x^2 + \Delta y^2} = \left\| \bm{\Delta t} \right\|$ and an absolute angle difference $\left| \Delta \theta \right|$ from a relative pose $\bm{\Delta \xi} = (\Delta x, \Delta y, \Delta \theta) = (\bm{\Delta t}, \Delta \theta)$, respectively.
Note that the operator $\ominus$ in Equations \ref{eq:accuracy-metric-trans} and \ref{eq:accuracy-metric-rot} computes a relative pose between two poses, i.e., $\bm{\xi}_j \ominus \bm{\xi}_i = \left( \bm{R}(\theta_i)^\top \left( \bm{t}_j - \bm{t}_i \right), \theta_j - \theta_i \right)$ for given $\bm{\xi}_i = (\bm{t}_i, \theta_i)$ and $\bm{\xi}_j = (\bm{t}_j, \theta_j)$.





\begin{table*}[h!]
    \centering \scriptsize
    \caption{Comparison of the trajectory errors in PF-SLAM, Graph-based SLAM, and Hector SLAM with and without the proposed FPGA accelerator.
    $\varepsilon_\mathrm{trans}$, $\varepsilon_\mathrm{rot}$, $\varepsilon_\mathrm{trans}^2$, and $\varepsilon_\mathrm{rot}^2$ are the accuracy metrics defined in Equations \ref{eq:accuracy-metric-trans} and \ref{eq:accuracy-metric-rot}.
    Graph-based SLAM offers a fast computation time, while at the same time achieving accuracy comparable to PF-SLAM.
    Hector SLAM does not produce a correct grid map of \textbf{ACES} dataset due to the absence of loop-closure detection.}
    \label{tbl:comparison-trajectory}
	\vspace*{5pt}
    \begin{tabular}{l|cccccc} \hline
        & \multicolumn{2}{c}{PF-SLAM} & \multicolumn{2}{c}{Graph-based SLAM}
        & \multicolumn{2}{c}{Hector SLAM} \\
        & CPU-only & CPU + FPGA
        & CPU-only & CPU + FPGA
        & CPU-only & CPU + FPGA \\ \hline
        \textbf{ACES} & & & & & & \\
        \quad $\varepsilon_\mathrm{trans}$ ($\mathrm{m}$)
        & $0.0785 \pm 0.1928$ & $\mathbf{0.0542 \pm 0.0569}$
        & $0.0571 \pm 0.0617$ & $0.0733 \pm 0.0996$
        & $0.4284 \pm 1.4093$ & $0.2292 \pm 0.7290$ \\
        \quad $\varepsilon_\mathrm{rot}$ ($\mathrm{rad}$)
        & $0.0409 \pm 0.0494$ & $\mathbf{0.0407 \pm 0.0492}$
        & $0.0410 \pm 0.0499$ & $0.0408 \pm 0.0495$
        & $0.0512 \pm 0.0714$ & $0.0485 \pm 0.0663$ \\
        \quad $\varepsilon_\mathrm{trans}^2$ ($\mathrm{m}^2$)
        & $0.0433 \pm 0.2988$ & $\mathbf{0.0062 \pm 0.0197}$
        & $0.0071 \pm 0.0242$ & $0.0152 \pm 0.0715$
        & $2.1696 \pm 9.2065$ & $0.5840 \pm 2.8771$ \\
        \quad $\varepsilon_\mathrm{rot}^2$ ($\mathrm{rad}^2$)
        & $\mathbf{0.0041 \pm 0.0116}$ & $0.0041 \pm 0.0121$
        & $0.0042 \pm 0.0124$ & $0.0041 \pm 0.0122$
        & $0.0077 \pm 0.0377$ & $0.0067 \pm 0.0331$ \\ \hline
        \textbf{Intel} & & & & & & \\
        \quad $\varepsilon_\mathrm{trans}$ ($\mathrm{m}$)
        & $0.1256 \pm 0.1292$ & $0.1117 \pm 0.0982$
        & $0.1223 \pm 0.1288$ & $0.1195 \pm 0.1148$
        & $0.1169 \pm 0.1140$ & $\mathbf{0.1076 \pm 0.1020}$ \\
        \quad $\varepsilon_\mathrm{rot}$ ($\mathrm{rad}$)
        & $0.0526 \pm 0.0736$ & $0.0504 \pm 0.0719$
        & $\mathbf{0.0499 \pm 0.0720}$ & $0.0504 \pm 0.0719$
        & $0.0555 \pm 0.0823$ & $0.0558 \pm 0.0817$ \\
        \quad $\varepsilon_\mathrm{trans}^2$ ($\mathrm{m}^2$)
        & $0.0325 \pm 0.0819$ & $\mathbf{0.0221 \pm 0.0425}$
        & $0.0316 \pm 0.1012$ & $0.0275 \pm 0.0718$
        & $0.0267 \pm 0.0649$ & $0.0220 \pm 0.0484$ \\
        \quad $\varepsilon_\mathrm{rot}^2$ ($\mathrm{rad}^2$)
        & $0.0082 \pm 0.0316$ & $\mathbf{0.0077 \pm 0.0298}$
        & $0.0077 \pm 0.0305$ & $0.0077 \pm 0.0304$
        & $0.0099 \pm 0.0338$ & $0.0098 \pm 0.0341$ \\ \hline
        \textbf{MIT-CSAIL} & & & & & & \\
        \quad $\varepsilon_\mathrm{trans}$ ($\mathrm{m}$)
        & $\mathbf{0.0379 \pm 0.0307}$ & $0.0386 \pm 0.0320$
        & $0.0480 \pm 0.0485$ & $0.0509 \pm 0.0703$
        & $0.0547 \pm 0.0713$ & $0.0469 \pm 0.0515$ \\
        \quad $\varepsilon_\mathrm{rot}$ ($\mathrm{rad}$)
        & $0.0172 \pm 0.0270$ & $\mathbf{0.0171 \pm 0.0256}$
        & $0.0383 \pm 0.0506$ & $0.0381 \pm 0.0501$
        & $0.0427 \pm 0.0495$ & $0.0368 \pm 0.0477$ \\
        \quad $\varepsilon_\mathrm{trans}^2$ ($\mathrm{m}^2$)
        & $\mathbf{0.0024 \pm 0.0059}$ & $0.0025 \pm 0.0068$
        & $0.0047 \pm 0.0132$ & $0.0075 \pm 0.0607$
        & $0.0081 \pm 0.0417$ & $0.0049 \pm 0.0215$ \\
        \quad $\varepsilon_\mathrm{rot}^2$ ($\mathrm{rad}^2$)
        & $0.0010 \pm 0.0048$ & $\mathbf{0.0009 \pm 0.0043}$
        & $0.0040 \pm 0.0128$ & $0.0040 \pm 0.0127$
        & $0.0043 \pm 0.0111$ & $0.0036 \pm 0.0118$ \\ \hline
    \end{tabular}
\end{table*}


\begin{table*}[h!]
    \centering \scriptsize
    \caption{Comparison of the trajectory errors between PF-SLAM with the proposed FPGA accelerator (16 particles) and other recent works, i.e., GMapping with FPGA accelerator for hill-climbing based scan matching (32 particles)~\cite{Sugiura21A}, GMapping (50 particles)~\cite{Kuemmerle09}, Google Cartographer~\cite{Hess16}, and Random Normal Matching~\cite{Ammon17}.}
    \label{tbl:comparison-trajectory-related-works}
	\vspace*{5pt}
    \begin{tabular}{l|ccccc} \hline
        & PF-SLAM (This work)
        & Sugiura \textit{et al.}~\cite{Sugiura21A}
        & K{\"u}mmerle \textit{et al.}~\cite{Kuemmerle09}
        & Hess \textit{et al.}~\cite{Hess16}
        & Ammon \textit{et al.}~\cite{Ammon17} \\
        & CPU + FPGA & CPU + FPGA & GMapping (50 particles)
        & Google Cartographer & Random Normal Matching \\ \hline
        \multirow{2}{*}{Environment} & \multicolumn{2}{c}{Pynq-Z2 (ARM Cortex-A9)}
        & N/A & Intel Xeon E5-1650 & Intel Core i7-4790 \\
        & \multicolumn{2}{c}{2 cores, 650MHz (CPU) + 100MHz (FPGA)} & N/A
        & 6 cores, 3.2GHz & 4 cores, 3.6GHz \\ \hline
        \textbf{ACES} & & & & \\
        \quad $\varepsilon_\mathrm{trans}$ ($\mathrm{m}$)
        & $0.0542 \pm 0.0569$ & $0.125 \pm 0.490$
        & $0.060 \pm 0.049$ & $0.0375 \pm 0.0426$ & N/A \\
        \quad $\varepsilon_\mathrm{rot}$ ($\mathrm{rad}$)
        & $0.0407 \pm 0.0492$ & $0.0852 \pm 0.319$
        & $0.021 \pm 0.023$ & $0.0065 \pm 0.0082$ & N/A \\ \hline
        \textbf{Intel} & & & & \\
        \quad $\varepsilon_\mathrm{trans}$ ($\mathrm{m}$)
        & $0.1117 \pm 0.0982$ & $0.117 \pm 0.130$
        & $0.070 \pm 0.083$ & $0.0229 \pm 0.0239$ & $0.0300 \pm 0.0576$ \\
        \quad $\varepsilon_\mathrm{rot}$ ($\mathrm{rad}$)
        & $0.0504 \pm 0.0719$ & $0.0859 \pm 0.284$
        & $0.052 \pm 0.093$ & $0.0079 \pm 0.0233$ & $0.0075 \pm 0.0064$ \\ \hline
        \textbf{MIT-CSAIL} & & & & \\
        \quad $\varepsilon_\mathrm{trans}$ ($\mathrm{m}$)
        & $0.0386 \pm 0.0320$ & $0.0505 \pm 0.0795$
        & $0.049 \pm 0.049$ & $0.0319 \pm 0.0363$ & $0.0646 \pm 0.0594$ \\
        \quad $\varepsilon_\mathrm{rot}$ ($\mathrm{rad}$)
        & $0.0171 \pm 0.0256$ & $0.0984 \pm 0.387$
        & $0.010 \pm 0.021$ & $0.0064 \pm 0.0064$ & $0.0107 \pm 0.0130$ \\ \hline
    \end{tabular}
\end{table*}


\begin{table}[h!]
    \centering
    \caption{Comparison of the trajectory errors between the original Hector SLAM with Gauss-Newton based scan matching, and Hector SLAM with the proposed FPGA accelerator.}
    \label{tbl:comparison-trajectory-hectorslam}
    \begin{tabular}{l|cc} \hline
        & \multirow{2}{*}{Gauss-Newton} & FPGA (CSM) \\
        & & + Gauss-Newton \\ \hline
        \textbf{ACES} & & \\
        \quad $\varepsilon_\mathrm{trans}$ ($\mathrm{m}$)
        & $0.3834 \pm 1.8286$ & $\mathbf{0.2292 \pm 0.7290}$ \\
        \quad $\varepsilon_\mathrm{rot}$ ($\mathrm{rad}$)
        & $0.0570 \pm 0.0901$ & $\mathbf{0.0485 \pm 0.0663}$ \\ \hline
        \textbf{Intel} & & \\
        \quad $\varepsilon_\mathrm{trans}$ ($\mathrm{m}$)
        & $2.1815 \pm 5.3097$ & $\mathbf{0.1076 \pm 0.1020}$ \\
        \quad $\varepsilon_\mathrm{rot}$ ($\mathrm{rad}$)
        & $0.2549 \pm 0.6629$ & $\mathbf{0.0558 \pm 0.0817}$ \\ \hline
        \textbf{MIT-CSAIL} & & \\
        \quad $\varepsilon_\mathrm{trans}$ ($\mathrm{m}$)
        & $0.0713 \pm 0.2002$ & $\mathbf{0.0469 \pm 0.0515}$ \\
        \quad $\varepsilon_\mathrm{rot}$ ($\mathrm{rad}$)
        & $0.0526 \pm 0.0981$ & $\mathbf{0.0368 \pm 0.0477}$ \\ \hline
    \end{tabular}
\end{table}

As shown in Table \ref{tbl:comparison-trajectory}, CSM core achieves accuracy close to the software implementation, which indicates that CSM core improves the performance without compromising the quality of outputs.
In some cases such as \textbf{ACES} in PF-SLAM and \textbf{MIT-CSAIL} in Hector SLAM, the accuracy of CSM cores even surpasses the one of the software counterpart.
PF-SLAM with FPGA acceleration achieves better accuracy than graph-based SLAM and Hector SLAM in all three datasets.
This is an expected result, as PF-SLAM maintains multiple hypotheses about the current state (i.e., robot trajectory and map) and selects the most probable one, which yields robustness and improved accuracy at the cost of increased computational costs (see Figures \ref{fig:gmapping-speedup} and \ref{fig:graphslam-frontend-speedup}).
Hector SLAM is not able to generate a globally-consistent map of \textbf{ACES} dataset due to large translational errors in robot poses; it suffers from the accumulated errors in robot poses, attributed to the absence of loop detection or any reliable mechanism to eliminate such errors during runtime.
Compared to this, PF-SLAM and graph-based SLAM detect loops at five intersections of corridors (see Figures \ref{fig:trajectory-comparisons} (center) and \ref{fig:grid-map-aces-sousoukan} (left)) and construct topologically correct grid maps.

We also quote the results from various studies along with the execution environment, i.e., our previous work~\cite{Sugiura21A}, GMapping with 50 particles~\cite{Kuemmerle09}, Google Cartographer~\cite{Hess16}, and Random Normal Matching~\cite{Ammon17}, in Table \ref{tbl:comparison-trajectory-related-works}.
CSM core offers better accuracy than our previous work~\cite{Sugiura21A}, where an FPGA-based accelerator for hill-climbing based scan matching is proposed and integrated to GMapping algorithm.
Since hill-climbing method does not converge to a correct solution unless a good starting point is given, our previous accelerator only works in the frontend scan matching, where the initial guess is close to the globally optimal pose.
In contrast, CSM core can handle backend loop detections with large initial errors as well as the frontend scan matching, which emphasizes the advantages of our proposal over \cite{Sugiura21A} on both robustness and versatility.
Our CSM core is able to produce results with the same quality as \cite{Kuemmerle09} using only one third the number of particles (50 and 16).
In \textbf{ACES} and \textbf{MIT-CSAIL} datasets, translational errors $\varepsilon_\mathrm{trans}$ are less than the twice of the grid map resolution $r$, which are comparable to the state-of-the-art methods~\cite{Hess16,Ammon17,Rodrigues21}.
Note that \cite{Hess16} and \cite{Ammon17} use a workstation with Intel Xeon E5-1650 (3.2GHz) and one with Intel Core i7-4790 (3.6GHz) to obtain these results, which are not the target computing platform in this paper.

Table \ref{tbl:comparison-trajectory-hectorslam} shows the comparison of trajectory errors between the original Hector SLAM and our improved version, which utilizes CSM cores to supply good starting estimates for Gauss-Newton scan matcher.
We can observe the significant improvement in accuracy: the original version fails to build globally consistent maps in all datasets, whereas the improved version performs successfully in \textbf{Intel} and \textbf{MIT-CSAIL} datasets.
As discussed in Section \ref{sec:design-application-to-hector-slam}, we set the number of resolution levels to three to compensate the weaknesses of gradient-based methods.

\subsection{Resource Utilization and Runtime Memory Consumption} \label{sec:eval-resource}
The FPGA resource utilization of our design (Figure \ref{fig:design-block-diagram}) is summarized in Table \ref{tbl:resource-utilization}.
The BRAM consumption is linear with the number of CSM cores implemented and also with the size of grid map buffers (Figure \ref{fig:design-block-diagram-core}), thus our design is constrained by the amount of BRAM available.
Thanks to the quantization of grid map values from 32-bit to 6-bit (i.e., 5.3x reduction in memory cost), grid map buffers only consume 68.6\% (96 slices) of BRAMs.

\begin{table}[htbp]
    \centering
    \caption{FPGA resource utilization of the CSM core}
    \label{tbl:resource-utilization}
    \begin{tabular}{l|rrrr} \hline
        & BRAM & DSP & FF & LUT \\ \hline
        Used & 111 & 24 & 20,121 & 21,026 \\
        Available & 140 & 220 & 106,400 & 53,200 \\
        Utilization (\%) & 79.29 & 10.91 & 18.91 & 39.52 \\ \hline
    \end{tabular}
\end{table}

In graph-based SLAM, the software scan matcher stores coarse maps to DRAM to avoid unnecessary computations for the same input.
This reduces the preprocessing cost when matching multiple scans to the same grid map (i.e., loop detections).
In contrast, CSM core computes coarse maps inside the sliding window maximum module (Figure \ref{fig:design-block-diagram-core}) and eliminates the necessity of DRAM storage for caching such coarse maps.
Using CSM core, the physical DRAM usage is reduced from 59.5 to 48.3 MiB (\textbf{ACES}), 101.7 to 76.6 MiB (\textbf{Intel}), and 60.1 to 39.5 MiB (\textbf{MIT-CSAIL}), respectively.

\subsection{Power Consumption} \label{sec:eval-power}
We measured the power consumption of the entire Pynq-Z2 board using a wattmeter: the power consumption was around 2.3-2.4W when we run three SLAM systems with and without our proposed CSM cores.
Note that the value reported above include the power consumption of CPU and other peripherals on the board as well as that of CSM cores, meaning that the CSM core itself consumes less than 2.3-2.4W.



\section{Conclusion} \label{sec:conc}
Despite the importance of SLAM on autonomous mobile robots, the stringent limitations of onboard computational resources hinders the use of SLAM on such robots.
To cope with high computational complexity and achieve real-time performance, a low-cost embedded device with a custom hardware accelerator is a promising computing platform.
In this paper, we proposed an FPGA-based universal accelerator compatible with a variety of 2D LiDAR SLAM methods.
We focused on the scan matching step as it becomes the major bottleneck, and implemented Correlative Scan Matching (CSM) on the programmable logic part, since CSM is both hardware-friendly and robust.
We conducted several architectural and algorithmic optimizations to fully exploit the inherent parallelism and reduce resource utilization.
Our design consisting of two CSM cores fits within a low-end FPGA (Pynq-Z2).

For comprehensive evaluations, we integrated the proposed accelerator into three representative SLAM approaches: scan matching, particle filter, and graph-based SLAM.
We confirmed that CSM core speeds up the scan matching by up to 14.09x, 14.84x, and 7.88x in these methods, and the loop-closure detection in graph-based SLAM by up to 18.92x, while only consuming 2.4W.
This reduced the wall-clock execution time by up to 4.67x, 4.00x, and 4.06x, enabling the real-time performance in typical indoor scenarios.
We qualitatively and quantitatively analyzed the outputs, and showed that CSM core produces accurate grid maps without distortion.
Results also highlighted the advantages of these methods: scan matching-based SLAM achieved the fastest computation time, graph-based SLAM was able to handle large environments by closing loops, and PF-SLAM provided more accurate results.
Our universal design allows to switch between these methods according to the environment conditions.
The error of trajectory estimate was less than $10 \mathrm{cm}$ and $0.05 \mathrm{rad}$ ($2.8^\circ$) in most cases, which was comparable to that of the software implementations and even the state-of-the-art SLAM methods.

\renewcommand{\baselinestretch}{1.0}
\bibliographystyle{unsrt}

\end{document}